\pdfoutput=1

\documentclass[11pt]{article}

\usepackage[preprint]{acl}

\usepackage{times}
\usepackage{latexsym}

\usepackage[T1]{fontenc}

\usepackage[utf8]{inputenc}

\usepackage{microtype}

\usepackage{inconsolata}

\usepackage{graphicx}

\usepackage{amsmath} 
\usepackage{cleveref} 
\usepackage{caption}

\usepackage{algorithm2e}
\usepackage{listings}

\definecolor{keywords}{RGB}{255,0,90}
\definecolor{comments}{RGB}{0,0,113}
\definecolor{red}{RGB}{160,0,0}
\definecolor{green}{RGB}{0,150,0}

\title{The Mystery of the Pathological Path-Star Task for Language Models}

\author{Arvid Frydenlund \\
  University of Toronto, Computer Science \\
  Vector Institute \\
  \texttt{arvie@cs.toronto.edu} 
  \\}

\usepackage[absolute,overlay]{textpos}

\begin{document}
\maketitle

\begin{textblock*}{18cm}(1.5cm,28.5cm) 
\centering
\footnotesize
   \href{https://aclanthology.org/2024.emnlp-main.695/}{Proceedings of the 2024 Conference on Empirical Methods in Natural Language Processing, pages 12493–12516 \\
   November 12-16, 2024 ©2024 Association for Computational Linguistics}
\end{textblock*}

\begin{abstract}

The recently introduced
path-star task is a minimal task designed to exemplify limitations to the abilities of language models \citep{bachmann2024the}.  It involves a {\em path-star} graph where multiple arms radiate from a single starting node and each node is unique.  Given the start node and a specified target node that ends 
an arm, the task is to generate the arm containing that target node.  This is straightforward for a human but surprisingly difficult for language models, which did not outperform the random baseline.
The authors hypothesized this is due to a deficiency in teacher-forcing and the next-token prediction paradigm.

We demonstrate the task is learnable using teacher-forcing in alternative settings and that the issue is partially due to representation.
We introduce a regularization method using structured samples of the same graph but with differing target nodes,  
improving results across a variety of model types. 
We provide RASP proofs showing the task is theoretically solvable. 
Finally, we find settings where an encoder-only model can consistently solve the task.  



\end{abstract}

\section{Introduction}

Language models (LMs) have become increasingly capable of solving a variety of complex tasks \citep{NEURIPS2020_1457c0d6, 52065, bubeck2023sparks}.  
LMs can do many spectacular things, making it surprising
when they perform poorly or require help on simple tasks  \citep{valmeekam2023on, wies2023subtask, golovneva2024reverse, berglund2024the, nezhurina2024alice}.  Recently, \citet{bachmann2024the} introduced one such seemingly simple task designed to showcase pathological behaviour of causal (decoder-only) autoregressive (AR) LMs trained via teacher-forcing. 
The task is simple by design and thus failure of AR models is both surprising and informative. We begin by describing the task in Sec.\@ \ref{sec:task}, before analyzing why it is difficult for LMs in Sec.\@ \ref{sec:methods}.



\subsection{The Path-Star Task}\label{sec:task}

We need to describe 
a path-star graph, $G$, i.e. the data meant to be manipulated, and its problem specification or question, $Q$, i.e. the prompt specifying the desired manipulation, 
and 
their tokenization. 

\begin{figure}
    \centering
    \includegraphics[scale=0.41,trim={0 0.6cm 0 0.5cm},clip]{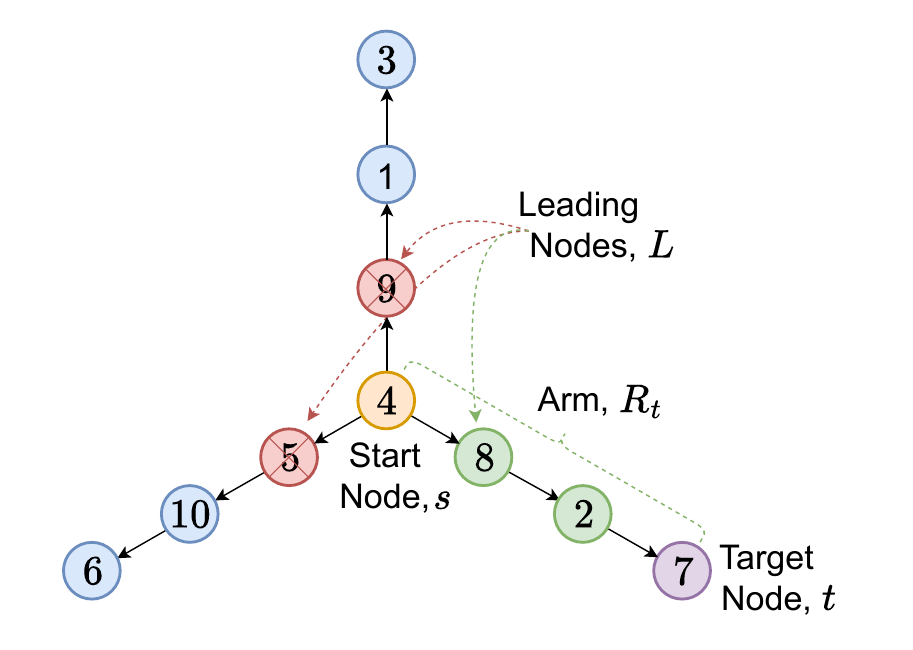}  
    \caption{An example path-star graph.  $D=3$, $M=4$, $s$ is `4', $t$ is `7', $R_t$ is `4 8 2 7', and $l_t$ is `8'.  One possible tokenization of $[G,\, Q,\, R_t]$, where {\bf the edges are permuted} is:  `BOS 9 1 | 10 6 | 8 2 | 2 7  | 1 3 | 4 8 | 4 5 | 5 10 | 4 9 | / 4 7 = 4 8 2 7 EOS'.
    One tokenization where the {\bf arms are permuted} is: `BOS 4 9 | 9 1 | 1 3 | 4 8 | 8 2 | 2 7 | 4 5 | 5 10 | 10 6 | / 4 7 = 4 8 2 7 EOS'.  
    }
    \label{fig:psg}
\end{figure}

Let $N$ be the set of unique nodes forming $G$.  A path-star graph contains one central starting node $s \in N$ and $D$ radial arms each of length $M$ (inclusive of $s$), s.t. $|N| = D(M-1) + 1$.  $s$ has degree $D$, all final nodes which end an arm, $F \subset N$ s.t. $|F| = D$, have degree 1, and all others have degree 2. See Fig.\@ \ref{fig:psg} for an example path-star graph.

Given $G$, and a task specification, $Q$, containing $s$ and a target node $t \in F$, the task is to generate the unique arm, $R_t$, as a sequence of nodes starting 
from $s$ until $t$.
i.e. $R_t = \mathrm{sort}(\{\,r  \in N \,|\, \forall_{f \in F}\, \mathrm{dist}(r,\, t) \leq \mathrm{dist}(r,\, f) \})$.\footnote{`dist' is graph distance. We abuse notation by treating $R_t$ as a set and $G$ and $Q$ as sequences after having been tokenized.}
Let $L$ be the set of possible leading nodes which are adjacent to $s$ i.e.\@ $L = \{l \in N \,|\, \mathrm{dist}(l,\, s) = 1\}$. 
The challenge 
is predicting the correct leading node $l_t \in L \cap R_t$ 
from all other leading nodes.    
By design, there is a uniform $1/D$ chance of this given only $G$.   Prediction above chance {\em should} be possible by inferring the correct target arm and thus $l_t$ given $t$ in $Q$.  
Note, as all nodes are unique, predicting all non-leading nodes is deterministic given their preceding neighbour 
(going from the direction of $s$ to $f \in F$).      

When generating the dataset, the nodes in a single graph are uniformly sampled from a set of possible nodes, $V$, without replacement.  $G$ is tokenized as a series of $D(M-1)$ edges where each edge is internally ordered by distance to $s$ and marked by the special token `|' so that a given edge, $(u,\, v)$, is a three token sequence `$u$ $v$ |'.  $Q$ is tokenized as a sequence of four tokens with special tokens marking the beginning and end of $Q$ as `/ $s$ $t$ ='.  Special beginning- and end-of-sequence tokens are also used, making the final vocabulary size $|V| + 5$.

\subsection{Autoregressive models and training}

A causal or decoder-only AR LM models the joint probability of a $T$-length sequence, $y$, as a factorized product of local probabilities, as in
\begin{equation}\label{eq::AR}  
p(y_1,\, y_2,\, \dots,\, y_{T} \,|\, y_0) = \prod\nolimits_{j=1}^{T} p(y_j\,|\, y_{< j}).    
\end{equation}  
Here, we model the path-star task as
\begin{equation}
    p(r_1,\, \dots,\, r_M\,|\, [G,\,Q]) = \prod_{j=1}^M p(r_j\,|\, [G,\,Q,\, r_{< j}]).
\end{equation}
Let $x=[G,\,Q,\,r_{< j}]$ be the concatenation of the tokenized graph and problem specification along with the partial ground-truth sequence, $r_{< j}$, which forms the given conditioning input to the model.  Such a model is trained via maximum likelihood, generally referred to as `teacher-forcing' in the context of language models, as the partial ground-truth sequence is used to condition the model during training instead of the model's own predictions as done during inference \citep{williams1989learning}.   We minimize $-\sum_{r \in R_t}\log p(\,r\,|\, x)$.  Thus the loss is only over the target sequence, $R_t$, and not on tokens in the prefix $[G,\,Q]$.  This is necessary as the nodes in $G$ are random and $Q$ 
must be provided, 
making them both unpredictable.  


$G$ and $Q$ are provided during inference. We evaluate using a non-autoregressive `teacher-forced' inference procedure 
that conditions on the partial ground-truths. 
 Thus the inference and training procedures match and prevent any potential training-inference bias (see Appx.\@ \ref{app:inference} for explanation).  

We focus on transformer models \citep{vaswani2017attention}, where the causal parameterization of AR models is enforced via an attention mask which prevents the token at any step $j$ from depending on any token at step $> j$.  This causal restriction applies across the entire input $x$.  
Positional embeddings make each token unique.  
To prevent learning a trivial answer based on position, as a data preprocessing step, {\bf  the edges are shuffled, which can be seen as a random permutation applied to the order of the edges in the tokenization of $G$.}

\subsection{Failure to learn:  Clever Hans hypothesis}\label{sec:failure}  

\citet{bachmann2024the} empirically demonstrated 
three different LMs -- finetuned GPT2, a smaller GPT2 trained from scratch \citep{radford2019language}, and a state-space model, Mamba \citep{gu2023mamba}
-- all fail to predict above $1/D$ chance, even in settings as small as $D=2$ and $M=5$.  They hypothesized 
this was 
caused by  
teacher-forcing.
This is due to there being two possible modes of prediction that the model can learn.  The first is the desired mode which 
learns to represent the entire path between $s$ to $t$. 
This mode is necessary for predicting $l_t$. 
Whereas, the second mode 
makes trivial predictions about the next node in the arm given the previous node.  This mode only needs to learn superficial information about individual edges but not the graph structure and is, by task design, 
sufficient for predicting all non-leading nodes given the correct preceding node.

\citet{bachmann2024the} argued that teacher-forcing will result in learning the second mode, referred to as the {\em Clever Hans} cheat (CHC).
This is because teacher-forcing conditions on the correct ground-truth, which in this case is the correct preceding node in the arm.  Also, when applied to AR models, it is restricted to making a single next-token prediction and hence precludes learning any long-term planning. 
Then, once the CHC is learnt, it will discourage learning the desired mode necessary for predicting $l_t$.  
Their intuition, which admittedly is not proven, is that sequence modelling relies on the intermediate training steps across the sequence to form a coherent representation of the overall sequence.  In our case, that would be a representation of the entire arm structure, however, here those intermediate steps do not participate in learning such a structure but are rather absorbed into learning the trivial CHC, resulting in a loss of this intermediate training signal.     

They presented empirical evidence for the CHC hypothesis by considering the overall sequence accuracy 
when provided with the correct preceding predictions. 
Here, all non-leading tokens are learnt with 100\% accuracy and the leading token is only predicted at $1/D$ chance, leading to an overall sequence accuracy of $1/D$ (See their Fig.\@ 3 and our Fig.\@ \ref{fig:CHC} in Appx.\@ \ref{app:CHC}). 
Interestingly, a trivial solution to the task exists if the model can look-ahead $M$ tokens to the end of the arm as the model just needs to find and match the correct target token.  Once done, it can apply the CHC in reverse order to determine the arm.  This led them to provide a supporting empirical argument, 
where they modified the task to require that the arm be generated in reverse order.  This makes the task trivial as the CHC can just be applied in reverse order via reverse supervision.   


Importantly, they established, a), the failure 
is not due to the amount of training data (see Sec.\@ \ref{sec:conta}), b), the failure is in-distribution, and 
c), 
it is not due to any exposure bias or other differences between the training and inference procedures \citep{bengio2015scheduled, ranzato2016sequence, arora-etal-2022-exposure}.  This allows them to dismiss these alternative explanations and conclude that the CHC causes the learning problem which is itself a consequence of teacher-forcing and next-token prediction.  

{\bf This leads to a discussion concerning possible fundamental limitations to the next-token prediction paradigm, with the path-star task being offered as a counter-example to the paradigm being sufficient to learn any task  
and also to conjecture that these limitations may apply to a broader set of more complex planning tasks}.

They then introduced a `teacher-less' model \citep{monea2023pass}, which uses $M$ masked tokens, $m$, to predict all tokens independently of the ground-truths i.e. $x=[G,\,Q,\,m_1,\,\dots,\,m_M]$. This eliminates teacher-forcing as it removes all input dependencies between target-side tokens.  Of the 15 reported experiments, this method allows the model to solve the task in 5 instances: for the $D=2$ experiment using the small GPT2, and for $D \in \{2,\,3,\,4\}$ (but not $D=5$) when using large GPT2.  Thus while the method did not work consistently, it acts as further empirical evidence that teacher-forcing is the issue and a potential alternative learning paradigm to next-token prediction.      









\section{Methods and Results}\label{sec:methods}

We solely focus on the small LM setting under the belief that such models should be able to learn such a simple task and that the biases from the pretrained data and any emergent abilities of LLMs will just obfuscate the root problem.  We implement our models using Fairseq \citep{ott2019fairseq}.  
Our AR models have $200$ 
dim.\@ embeddings and 6 layers, each with $800$ dim.\@ feed-forward projections and 8 
heads. This is smaller than both 12- and 36-layer
GPT2 models used by \citet{bachmann2024the}.  
We use Adam 
with a learning rate of 0.0005, a dropout rate of 0.1, and a weight decay of 0.01.  We use 16-bit training 
and a batch size of 1024. Each model is given a maximum of 100 epochs but stops if the validation loss drops below 0.001. 
Each experiment is trained from scratch.

Following \citet{bachmann2024the}, $|V|=100$ and $M=5$. Each dataset for $D \in \{2,\,3,\,4,\,5\}$ is made up of 2,000,000 training and 20,000 test samples of randomly generated $(G,\, Q)$ pairs without overlap. Unlike them, we randomly permute $G$ at every epoch instead of just once prior to training, in an attempt to prevent overfitting.   

We present our work as an investigation over a series of hypotheses and corresponding experiments to get at the heart of the path-star mystery.  As such, we report intermediate results and describe new methodology as it becomes motivated. 
We present results in order of our findings, however, we need to give some post-hoc explanations for our methodology 
to inform the contents of Tables \ref{tab:decoder-only}, \ref{tab:encoder-decoder-ar}, \ref{tab:encoder-decoder-iar},  \ref{tab:encoder-only}.  
First, in initial experimentation (under modified task conditions), 
we found that the models could solve the task seemingly at random.  This motivated the use of running multiple seeded trials for each experiment. 
For all experiments, we consider the percentage of trials that correctly succeed in learning the task across 11 trials.  Second, we found it necessary to set attention dropout to zero, 
since the task requires routing exact node information. Third, we also found that we required learned- instead of sinusoidal-positional-embeddings.  We suspect that the latter results in too strong of a positional bias when 
permuting the edges in $G$.  
We also found no 
embeddings works for the decoder-only model as positional information will arise out of the asymmetry induced by causal masking \citep{tsai-etal-2019-transformer, haviv-etal-2022-transformer, NEURIPS2023_4e85362c}

\subsection{A reproduction of empirical results}

\begin{table*}[t]
    \centering
    \begin{tabular}{c|c|c|c|c|rr|rr|rr|rr}
       ID & Perm.  & $Q$  & Tgt./Dir. & S.  & \multicolumn{2}{c|}{$D=2$}  & \multicolumn{2}{c|}{$D=3$}  & \multicolumn{2}{c|}{$D=4$}  & \multicolumn{2}{c}{$D=5$} \\ \hline

       1 & Edge & End & Fwd. & 0 & 0\% & 0\%  & 0\%  & 0\%  & 0\%  & 0\%  & 0\%  & 0\%  \\  
       2 & Edge & End & Rev. & 0 & 100\% &  & 100\%  &  & 100\%  &  & 100\%  &  \\  
       3 & Edge & End & $l_t$-only & 0 & 0\% & 0\%  & 0\%  & 0\%  & 0\%  & 0\%  & 0\%  & 0\%  \\  \hline 
       4 & Arm & End & Fwd. & 0 & 100\%  & & 36\%  & 0\%  & 9\%  & 0\%  & 9\%  & 0\%  \\  
       5 & Arm & Start & Fwd. & 0 & 100\% &  & 100\%  &  & 100\%  &  & 100\%  &  \\ 
       6 & Edge & Start & Fwd. & 0  & 0\% & 0\%  & 0\%  & 0\%  & 0\%  & 0\%  & 0\%  & 0\%  \\ \hline  
    
       7 & Arm & End & Fwd.& 1 & 100\% &  & 91\%  &  9\% & 91\%  & 9\%  & 36\%  & 55\%  \\  


       8 & Edge & End & Fwd. & 1 & 0\% &  0\% & 0\%  & 100\%  & 0\%  &  100\% & 0\%  & 100\%  \\  
       9 & Edge & Start & Fwd. & 1 & 0\% &  0\% & 0\%  & 100\%  & 0\%  &  100\% & 0\%  & 100\%  \\  
       10 & Edge & Start & Fwd. & 2 & \multicolumn{2}{c|}{NA}   & 0\%  & 0\%  & 0\%  &  91\% & 0\%  & 100\%  \\  \hline
       9x & Edge & Start & Fwd. & 1 & 0\% &  0\% & 0\%  & 0\%  & 0\%  &  0\% & 0\%  & 100\%

    \end{tabular}
    \caption{Percent of successful trials ($n$=11) using the AR (decoder-only) model.  `ID' is the exp.\@ ID. 
    `Perm.' is the type of 
    permutation applied to $G$ (Sec.\@ \ref{sec:perm}).  `$Q$' is the relative position of $Q$ to $G$ 
    (Sec.\@ \ref{sec:pos}).  `Tgt./Dir.' is the type of target we are trying to generate (Sec.\@ \ref{sec:against}).  And `S.' is the number of structured samples used (Sec. \ref{sec:conta}).    
    For each exp. in $D \in \{2,\,3,\,4,\,5\}$, we report the percent of 
    trials 
    which learnt the task to at least 
    95\% sequence accuracy 
    in the first column.  In the second column, we report the percent of unsuccessful trials where the valid and training loss has {\em not} diverged i.e. 0\% means all trials have overfit (Sec.\@ \ref{sec:conta}). `x' IDs use a larger model (Sec.\@ \ref{sec:rasp}).}     
    \label{tab:decoder-only}
\end{table*}

As the results of \citet{bachmann2024the} are surprising, we independently verified them as an initial step.  Experiment (exp.) 1 of Table \ref{tab:decoder-only}, confirms that the task is not learnable under the initial conditions.
Exp.\@ 2 confirms that reversing the arm results in a trivial 100\% success rate.

\subsection{Simplifying the task}

Having confirmed the results in the original task setting, our method to investigate the issue will be to simplify the task until it becomes consistently solvable.  We begin by considering the target-side.

\subsubsection{Isolating the CHC from the task}\label{sec:against}

In Sec.\@ \ref{sec:failure}, we indicated the CHC causes the learning problem.  We first consider a simplified version of the task where only $l_t$ is predicted instead of the entire arm, $R_t$.\footnote{Our work considered a preprint version of \citet{bachmann2024the}.  In the final version, they also conducted this exp. (their Appx.\@ F.5).  This version, along with correspondence with an author, clarified their position that the presence of the CHC does not cause the learning issues directly.  Thus when they argued that after learning the CHC, `it is significantly harder for the model to learn the correct solution now', they are considering a hypothetical point in training when the model has access to $R_t$ and could potentially learn $l_t$ before the CHC is learnt and absorbs the critical information in $R_t$ and also prevents its recovery via alternative learning paths.}  Exp.\@ 3 of Table \ref{tab:decoder-only}, shows 
this 
produces the same negative result as when predicting the entire arm.  This allows us to exclude the possibility that $l_t$ is indecipherable due to the model being overwhelmed by the CHC i.e. that the presence of the CHC causes the learning issues directly.

Having clarified this, we can state the CHC hypothesis as being that teacher-forcing will cause the CHC to be learnt, which will absorb critical supervisory information in $R_t$ needed for learning the task.  There are underlying assumptions that $R_t$ is both necessary\footnote{{\em Necessary} may be too strong of a statement since \citet{bachmann2024the} stated that the CHC makes it `harder or potentially intractable to learn the true mechanism from the remaining tokens alone'.  But they also stated that $l_t$ `may become impossible to learn since the model is deprived of all information about the subsequent targets' and use the term `indecipherable' in reference to $l_t$.} and sufficient supervision to learn the task and that the task is inherently difficult without it.  As the evidence for the CHC hypothesis is empirical,  the degree of difficulty induced by lack of supervision rests on the empirical results, with the current results of not being able to predict above chance suggesting a high degree of difficulty.  

Exp.\@ 3 is defined so that the only 
supervision about the correct arm is $t$ via $Q$.  This begs the question, why is the task difficult only given $t$?  To solve the task, all the model needs to do is 1) determine all final nodes, 2) trace back each arm from the final node to leading node, and 3) predict the leading node matching with $t$.  Importantly, this requires that the model can represent the arms in the graph. This prompts the research questions of why is $t$ insufficient information to solve the task and what makes correctly representing the graph so challenging?  Note the CHC is a red-herring to such questions since, as discussed above, it just results in a situation where $t$ is the only given supervision.

As these questions stem from underlying assumptions of the CHC hypothesis, exploring them will be informative as to what impact the path-star task has on the next-token 
paradigm via teacher-forcing and planning tasks in general.  These questions motivate us to experiment with modifying the source-side graph representation of the task.\footnote{ 
All proceeding exps.\@ predict $R_t$ as in the original 
task.}  




\subsubsection{Alternative hypothesis: representation issues due to the permutation of $G$} \label{sec:perm}

Our first hypothesis as to what is preventing learning the solution is that it is a representation issue due to randomly permuting the edges of $G$.  This will corrupt the arm structure with the model seemingly unable to recover the structure.  In particular, when using a causal model, all information can only be routed forward in `time' and this may induce difficulties when trying to recover and route information across the arm structure.\footnote{See Appx.\@ \ref{appx:RASP} for discussion and proof routing is possible.}  
 Not only does edge-wise permutation make routing information across the arms harder, or even impossible, but the difficulty in learning might also be due to the assumptions we make when we decompose the joint probability as in Eq.\@ \ref{eq::AR}.  Specifically, we are parameterizing the model to a specific decomposition \citep{yang2019xlnet, liao-etal-2020-probabilistically}.  However, by permuting the edges, we are forcing the model to learn an exponential number of possible decompositions. This may be a challenge, even when using an over-parameterized model like a transformer, and may explain the difficulty of the task. 
 
Thus we can simplify the task where 
we retain the arm structure 
by only permuting the order of the arms relative to each other (but not the internal order of the edges). 
Refer to this change as {\em Edge}- vs. {\em Arm}-wise permutation (see Fig.\@ \ref{fig:data} in Appx.\@ \ref{appx:data}).   If this is solvable, then we know that the issue lies in the corruption of the arm information via permuting the edges.  Exp.\@ 4 of Table \ref{tab:decoder-only} shows this improves the results,  with $D=2$ being consistently solved, but with a diminishing success rate as $D$ increases.  These partial improvements lead us to a related hypothesis.

\subsubsection{Alternative hypothesis: representation issues due to the order of $G$ and $Q$}\label{sec:pos}

If we can only route information into the future, maybe our representation issue stems from the fact that we have placed the problem specification after the graph during tokenanization.  That is, we have placed the information needed to specify what to do with the data after the actual data \citep{chen2024premise}.  Thus the latent representation of $G$ formed by the LM can not depend on  $Q$.  Thus instead we form our input as $x=[Q,\,G,\, r_{< j}]$.  Refer to this as $Q$'s position being either {\em Start} vs. {\em End}.   Exp.\@ 5 of Table \ref{tab:decoder-only} demonstrates that this consistently solves the task when combined with permuting the arms only, but goes back to being completely unsolved when combined with permuting the edges (Exp.\@ 6).  

These last exps.\@ demonstrate that the task is solvable in alternative settings and that the representation of $G$ and $Q$ are aspects of the task's underlying difficulty.  Arm-wise permutation significantly decreases the difficulty of the task.  This is because it allows the graph to retain higher-order structural information, whereas edge-wise permutation eliminates all such information except for individual edges.  Thus edge-wise permutation is a culprit in the difficulty.  While this shows that the task is solvable, it is unsatisfying as we require stronger supervisory information in this setting.  
Since the causal constraint of decoder-only models potentially induces some of these issues, we are motivated to change the model specification to see if abandoning this constraint will solve the task.  

\subsection{Changing the model parameterization}

\subsubsection{Encoder-decoder model, or, alternative hypothesis: it's the causal constraint}

Here we use an encoder-decoder model 
with 
6 and 3 layers respectively and tied embeddings. 
Removing the causal constraint on the source-side encoding of $[G,\,Q]$ makes the 
position of $Q$ to $G$ irrelevant.  If this model can consistently solve the task, it will show that 
it is the causal constraint which prevents the decoder-only model from recovering the arm structure with edge-wise permutation.



\begin{table*}[t]
    \centering
    \begin{tabular}{c|c|c|c|rr|rr|rr|rr}
       ID & Perm.  & Dir. & S.  & \multicolumn{2}{c|}{$D=2$}  & \multicolumn{2}{c|}{$D=3$}  & \multicolumn{2}{c|}{$D=4$}  & \multicolumn{2}{c}{$D=5$} \\ \hline

       11 & Arm  & Fwd. & 0 & 100\% &  & 100\%  &  & 100\%  &  & 100\%  & \\  
       12 & Edge & Fwd. & 0 & 0\% & 0\%  & 0\%  & 0\%  & 0\%  & 0\%  & 0\%  & 0\%  \\ \hline  
       
       13 & Arm & Fwd. & 1 &  100\% &  & 100\%  &  & 100\%  &  & 100\%  &   \\  

       14 & Edge & Fwd. & 1 & 9\% & 0\%  & 27\%  & 9\%   & 0\%  & 9\%   & 0\%  & 100\%  \\  
       15 & Edge & Fwd. & 2 &  \multicolumn{2}{c|}{NA}  & 45\%  & 0\%   & 0\%  & 9\%   & 0\%  & 73\%  \\ \hline  
       14x & Edge & Fwd. & 1 & 0\% & 0\%  & 0\%  & 0\%   & 10\%*  & 0\%   & 0\%  & 0\% 

    \end{tabular}
    \caption{Results using the encoder-decoder AR model. 
    *Only 10 trials completed.}  
    \label{tab:encoder-decoder-ar}
\end{table*}

Exp.\@ 12 in Table \ref{tab:encoder-decoder-ar} shows that using a non-causal encoder 
does not solve the task with edge-wise permutation. This motivates us to revisit the `teacher-less' methodology as it has been shown to partially work and is an alternative non-causal methodology in Sec.\@ \ref{sec:IAR}.  
While this experimental setup does not learn the task as is, we will revisit it in Sec.\@ \ref{sec:conta} and show that it is possible (Exp.\@ 14 and Exp.\@ 15).

\subsubsection{Non-autoregressive models}\label{sec:IAR}


\begin{table*}[t]
    \centering
    \begin{tabular}{c|c|c|c|c|rr|rr|rr|rr}
       ID & Perm.  & Train  & Dir. & S.  & \multicolumn{2}{c|}{$D=2$}  & \multicolumn{2}{c|}{$D=3$}  & \multicolumn{2}{c|}{$D=4$}  & \multicolumn{2}{c}{$D=5$} \\ \hline

       16 & Arm & IAR & Fwd. & 0 & 100\% &  & 100\%  &  & 82\%  &  0\%  & 82\%  & 0\%  \\  
       17 & Edge & IAR & Fwd. & 0 & 0\% & 0\%  & 0\%  & 0\%  & 0\%  & 0\%  & 0\%  & 0\%  \\ \hline  
       
       18 & Arm & IAR & Fwd. & 1 &  100\% &  & 100\%  &  & 100\%  &  & 100\%  &   \\  

       19 & Edge & IAR & Fwd. & 1 & 64\%  & 0\%   & 9\%  & 0\%   & 0\%  & 0\%  & 0\% & 0\%   \\   
       20 & Edge & IAR & Fwd. & 2 &  \multicolumn{2}{c|}{NA}   & 18\%  & 0\%   & 0\%  & 100\%  & 0\% & 100\%   \\ \hline  
       19x & Edge & IAR & Fwd. & 1 & 0\%  & 0\%   & 0\%  & 0\%   & 0\%  & 36\%  & 0\% & 91\%   

    \end{tabular}
    \caption{Results using the CMLM (encoder-encoder) IAR model with IAR training (teacher-forcing).}     
    \label{tab:encoder-decoder-iar}
\end{table*}

\begin{table*}[t]
    \centering
    \begin{tabular}{c|c|c|c|c|rr|rr|rr|rr}
       ID & Perm.  & Train   & Dir. & S.  & \multicolumn{2}{c|}{$D=2$}  & \multicolumn{2}{c|}{$D=3$}  & \multicolumn{2}{c|}{$D=4$}  & \multicolumn{2}{c}{$D=5$} \\ \hline

       21 & Arm & IAR & Fwd. & 0 & 100\% &  & 82\%  & 0\%  & 36\%  & 0\%  & 9\%  & 0\%  \\  
       22 & Edge & IAR & Fwd. & 0 & 36\% & 0\%  & 0\%  & 0\%  & 0\%  & 0\%  & 0\%  & 0\%  \\ \hline  
       
       23 & Arm & IAR & Fwd. & 1 &  100\% &  & 100\%  &  & 100\%  &  & 91\%  & 9\%   \\  

       24 & Edge & IAR & Fwd. & 1 & 100\% &  & 45\%  & 55\%  & 18\%  &  82\%  & 9\%  &  91\% \\   
       25 & Edge & IAR & Fwd. & 2 &  \multicolumn{2}{c|}{NA}   & 55\%  & 45\%  & 36\%  &  64\%  & 11\%*  &  89\% \\ \hline  

       24x & Edge & IAR & Fwd. & 1 & 100\% &  & 100\%  &  & 91\%  &  0\%  & 100\%  & \\
       22x & Edge & IAR & Fwd. & 0 & 100\% &  & 55\%  & 0\%  & 73\%  & 0\%  & 18\%  & 0\%  \\ \hline
       26x & Edge & NAR & Fwd. & 1 & 100\% &  & 100\%  &  & 91\% &   9\%  & 64\%  &  36\%  \\

    \end{tabular}
    \caption{Results using the encoder-only IAR model.  *Only 9 trials were completed for this experiment.}     
    \label{tab:encoder-only}
\end{table*}

\citet{bachmann2024the} reported that `teacher-less' models were unable to solve the task in the small LM  setting and we attempt to improve their results.  However, we alter their `teacher-less' model as it was designed to modify a pre-trained LM post-hoc and so is inappropriate to our settings   \citep{monea2023pass}.   Rather, we note it is actually a kind of non-autoregressive model (NAR)  \citep{gu2018non, wang-etal-2018-semi-autoregressive, gu-kong-2021-fully}.  

NAR models treat all targets as independent in order to make multiple predictions in parallel instead of sequentially. This is achieved by removing the causal constraint i.e.\@ attention mask.  Complete independence is assumed for (fully) NAR models.   However, this can lead to poor performance as it limits the ability to learn from dependencies inherent in the sequence \citep{lee-etal-2018-deterministic, qian-etal-2021-glancing}.  This led to the development of iterative autoregressive models (IAR)\footnote{Often called iterative NAR models, which is a misnomer.} which assume partial dependencies, both during training and inference -- except in the first generation step \citep{lee-etal-2018-deterministic, ghazvininejad-etal-2019-mask}.  Importantly, IAR models assume no order-of-generation,  
which may allow them to learn the reverse-order solution without specific reverse-order supervision.

To train an IAR model, an order-permutation of the targets, $R_t$, is sampled, along with a time-step, $j$ such that model conditions on the permuted or `unmasked' ground-truths prior to step $j$.  This is equivalent to the MLM objective with a dynamically sampled masking rate, where the uniform masking acts as the permutation \citep{devlin-etal-2019-bert, lee-etal-2018-deterministic, ghazvininejad-etal-2019-mask}.  Thus 
IAR models 
use teacher-forcing, but applied to a permuted target sequence.  We use CMLM  \citep{ghazvininejad-etal-2019-mask} for an encoder-encoder IAR model as well as an encoder-only model using the same hyper-parameters as the decoder-only model i.e.\@ similar 
but without causal masking.  

Both NAR and IAR models make use of the full {\em target} supervision of $R_t$ by bypassing the CHC.  For NAR models, this is achieved by removing all {\em input} supervision and hence teacher-forcing.  For IAR models, this is achieved via input samples that mask non-contiguous tokens in $R_t$, thus excluding the trivial edge-lookup shortcut employed by CHC.




We evaluate the IAR models using both 1-step NAR and $M$-step IAR inference 
with both producing the same results (thus we only report one).  That is, once, the model learnt the solution, it could generate the entire arm in one step just as well as over $M$ steps.  In principle, the IAR models should be able to first generate $t$ in the last position, condition on it, and then just generate the arm in reverse order via CHC -- which should be much easier than learning the true solution.  This did not happen.\footnote{It should be disconcerting for practitioners of IAR models that the trivial generation order does not seem to be found. However, this may be an artifact of the task's underlying difficulty and not an issue with IAR models in general.}    

We demonstrate that small IAR models are capable of learning the arm-wise task in Tables \ref{tab:encoder-decoder-iar} and \ref{tab:encoder-only}.  The encoder-only model is the first model to have successful trials on the edge-wise task (Exp.\@ 22).  Exps.\@ 16 and 21 vs. 5 and 13  suggest IAR models may not be as performative, 
however, this does not hold once structured samples are used (Sec.\@ \ref{sec:conta}).

\subsection{Sensitivity and structured samples}\label{sec:conta}

Many of the arm-wise exps.\@ across models and the encoder-only edge-wise exp.\@ had failed trials but also successful ones.  
Observing the training and validation losses of these failed trials led us to conclude that the model was overfitting in these trials. 
Consider the first plot of Fig \ref{fig:contra} (in Appx.\@ \ref{appx:samples}) with $D=2$.  Here all trials successfully learn the task, which can be seen when the validation accuracy branches off from chance.  However, the last trial nearly overfits.  The third plot has the same 
setup but with $D=4$.  Here only a single trial succeeded and the rest maintained a stagnant validation accuracy at chance while the training and validation losses diverged.   To prevent this overfitting, we experimented with standard regulatory methods such as lowering the learning rate, increasing batch size and L2 regularization, etc.\@ without success.

This 
leads us to reconsider what may cause overfitting. In the edge-wise task, even though the edges are shuffled every 
epoch, there is only a set number of graphs and targets. 
Taking a step back, when we consider the task after the CHC has been learnt, the model needs to be able to identify $l_t$ solely via $t$.  Thus the solution depends on or is sensitive to a single token, $t$.
As all nodes are unique, the space of sampled graphs 
to targets is large with    
\begin{equation}
    Z = \frac{|V|!}{(|V| - D(M-1) -1)!} \times D  
\end{equation}
possible input-target pairs.  This implies that the dataset will contain nearly all uniquely sampled graphs to any target i.e. we will not encounter the same graph but with different target arms.  However, because of this, the models can learn many possible spurious correlations between any node or combination of nodes in $G$ to $t$ and 
easily 
overfit.

To prevent this, we develop a method where we supplement the training data with multiple instances of a single $G$ but paired with different targets, and hence, different $Q$ and target arms. 
This is achieved via expanding each batch with one or more of these  {\em structured samples} per original $G$. 
These extra instances should act as interference on any spurious training signal.  Thus these samples do not reduce the sensitivity but rather they inform the model that the task is sensitive to $t$ in particular.  
Hence, this can be viewed as extra supervisory information applied at the batch-level. 


We now revisit prior experiments with structured samples (`S.' \@ $> 0$ in all tables).  
For the decoder-only model, we find this reduces overfitting as intended but does not lead to successful trials.
Exp.\@ 7 (arm-wise, $Q$-end) shows improved success rates at $D > 2$.  This can be observed in the second and fourth plots in Fig.\@ \ref{fig:contra} 
where the validation loss now tracks the training loss when provided with structured samples
in contrast to their corresponding first and third plots with diverging losses. 
It also leads to learning the solutions in fewer epochs in the $D=2$ case and leads to 10/11 instead of 1/11 of the trials succeeding in the $D=4$ case.  

 For all other models, this method improves success rates across all edge-wise experiments (Exps.\@ 14, 15, 19, 20, 24, and 25).  Exps.\@ 14 and 15 show the first successes with the encoder-decoder model.  {\bf  Despite being inconsistent, this is an important result, as    
it demonstrates that casual models can learn the task via teacher-forcing.}  As the CHC will reduce the supervision to $t$, it also shows that $t$ may be sufficient supervision 
to learn the task with next-token prediction.  Thus this 
result challenges the empirically supported CHC hypothesis.  

While these are strong improvements, we have not solved the mystery as the encoder-only model with $D=2$ is the only consistently solved setting (Exp.\@ 24). Exps.\@ 15, 20, and 25 show that increasing the number of samples helps but has diminishing returns.  These results indicate that overfitting is a culprit to the task's difficulty, but not the only one. 
The decoder-only model has no successful trials, but the encoder-decoder model has a few.  Also, the encoder-only model outperforms the  
encoder-encoder model (Exps.\@ 24 and 25 vs.\@ 19 and 20).  The only difference between these last two 
is parameterization and not the training method. 
Thus 
the differences 
between each model's performance provide further evidence the task's difficulty depends on specific representations.       
Next, we turn to RASP to verify if the task is actually solvable for decoder-only models and to potentially gain insights into why the task is difficult just given $t$.


\subsection{RASP}\label{sec:rasp}

The RASP 
programming language is a formal computation model used to verify if a transformer is capable of solving a given symbolic (non-numerical) task where the existence of a valid RASP program 
proves there exists at least one transformer that can solve the task \citep{weiss2021thinking, zhou2024what}.  Note this does not imply that such a transformer is learnable but only that it can represent the task. 
A RASP program transforms a 
sequence of tokens into a new one 
using operations that can be implemented via transformers, such as  
element-wise operations which mirror feed-forward layers and `select' and `aggregate' operations which combine to mirror the attention mechanism. 
See Appx.\,\ref{appx:RASP} for details.  Each operation corresponds to a transformer layer.  As such, loops are disallowed and the length of a RASP program upperbounds the task's difficulty 
in terms of the required number of layers. 



We develop four 
algorithms for solving the task with edge-wise permutation using non-causal encoders
(Listings are in Appx.\@ \ref{appx:RASP-path-star}), proving there exists multiple 
transformers which can solve the task. 
Three of these, Listings \ref{func:non-causal-propagate-back-target}, \ref{func:non-causal-propagate-backward-targets}, and \ref{func:non-causal-propagate-forward-start}, work by traversing across the arms (in parallel) one edge at a time.  For example, Listing \ref{func:non-causal-propagate-backward-targets}, routes the identity of each arms' final node, 
starting at each final node and moving backward 
until reaching the corresponding leading nodes.  This results in a sequence where certain edges now contain pairs of corresponding leading and final nodes, $(l_d,\, f_d)$, which can be used to match $f_d = t$ to identify $l_d = l_t$.  

Importantly, these all require loops over 
$M$, which is invalid in general, but valid in our case, as $M$ is a static value. However, each step in the traversal requires one or more transformer layers to implement, and thus 
require $\mathcal{O}(M)$ number of layers.  Listing \ref{func:non-causal-propagate-log} provides a $\mathcal{O}(\log M)$ algorithm,\footnote{Incidentally, this uses the same {\em doubling} parallel graph traversal algorithm used by \citet{pmlr-v235-sanford24a} to prove the $k$-hops task is solvable in $\mathcal{O}(\log k)$.  Interestingly, they find transformers can actually learn the log algorithm for $k$-hops.}  provided we have $\mathcal{O}(M)$ extra sequence tokens to store intermediate results (the edge markers '|' can be used).  However, this algorithm is more complex (from a human perspective).  Despite the empirical results, we prove that a decoder-only model can solve the task (Listing \ref{func:causal-propagate}).  This algorithm is complex as it can only route information into future edges and thus requires different rules 
whether connecting edges come before or after a current edge.  {\bf This potentially explains why the task may be harder to learn for decoder-only models}.   

RASP can explain why the arm-wise permutation is an easier task.  Listing   \ref{func:non-causal-arms-propagate-backward-targets} describes a $\mathcal{O}(1)$ algorithm which works by just finding the edges with a final token and then using positional information to `jump' to the corresponding leading nodes.  

Thus the main results of this analysis are: 1) the task is solvable by transformers, 2) multiple solutions exist, 3) the `simplest' algorithms require a linear number of layers to $M$, and  4), at least one causal transformer can, provably, solve the task. 
The first three of these results give an actionable insight since we can increase the number of layers. 
This is antithetical to the overfitting problem, however, it may be that as we increase 
layers, more solvable algorithms become viable and may be easier to find during training.  Exp.\@ 24x shows this makes the task consistently solvable for the encoder-only model (with structured samples, 
Exp.\@ 24x vs. 22x), but all other models do worse in success rate and/or overfitting (Exps.\@ 9x, 14x, 19x).

\section{Conclusion}\label{sec:conclusion} 

We finally have a model that consistently solves the task, however, we still have many open questions. 
In this section, we consider these questions as well as the progress we have made in the task.  

First, we consider the role sensitivity has in the task's difficulty.  In Sec.\@ \ref{sec:conta} we hypothesized the model's overfitting was due to sensitivity \citep{hahn2021sensitivity, chen-etal-2023-relation, chakraborty-etal-2023-zero}. 
In particular, we argued that the solution is sensitive to a single token and that the model needs to learn this in order to solve the task.  However, because of the size of the sample space of graphs to targets, $Z$, and the fact that all nodes are unique, there will be many possible spurious correlations between $G$ and $t$ which the model can learn and solve the task via these shortcuts.  As these shortcuts are non-generalizable, the model will fail during inference.  Thus by using structured samples, we are exposing the underlying sensitivity of the task to the model by removing noise that would otherwise disguise this aspect of the task.  The empirical results show this accounts for some of the task's difficulty but not all of it. 

There is another role that sensitivity may play. 
When we consider the question of why $t$ is insufficient information, 
one reason may be that sensitivity to $t$ is the actual {\em cause} of the difficulty.  This is a conjecture the task is difficult {\em because} it is sensitive. A function is sensitive if small changes in the input cause large changes in the output.  
Sensitive functions are known to be hard 
for transformers to learn \citep{hahn2021sensitivity, bhattamishra-etal-2023-simplicity, hahn2024sensitive}. 
Such analyses depend on a formal description of the sensitivity of the task and we found formulizing this to be challenging.  

We ran an initial exp.\@ to evaluate this conjecture.  
Teacher-forcing should decrease sensitivity via the provided ground-truth (input) support.  Teacher-forcing is not viable for AR models.  However, IAR models have a viable teacher-forcing training method while NAR models foregoe teacher-forcing.  Thus finding a performance difference between them would empirically support this conjecture.   

We train a NAR model in Exp.\@ 26x to contrast with Exp. 24x.  Contrary to expectation, the performance of the NAR model is nearly as good as the IAR model, with $D=5$ showing the only difference.  We believe this result may not hold under increased task difficulty, especially as $M$ grows, since more samples can offer ground-truth support while bypassing the CHC.  As we have not formally defined sensitivity, this remains a conjecture only, with, potentially, some primary counter-evidence.   

One open question is why the task's difficulty increases with $D$ even though RASP analysis indicates the solutions do not depend on $D$.  This may be explained by $Z$ depending on $D$. It may also relate to the sensitivity conjecture, as 
$D$ increases the number of possible targets for a given graph. 

Another open question is why increasing the layers helps for the encoder-only model.  As there were some successful trials with the smaller model, it can not be the case that the number of layers was a constraint on solving the task, however, it seems easier to find the solution with more layers.

Perhaps the biggest open question is why only the encoder-only model consistently solves the task.  If this is solely due to the non-causal parameterization, we would expect the encoder-encoder model to perform the same as the training method is identical (Exps.\@ 19, 20, 19x vs.\@ 24, 25, 24x).  As this is not the case, there must be a significant difference between these two non-causal parameterizations. This was an unexpected finding.  The encoder-only model is unique in that the source-side representation conditions and is dependent on the target-side.  This is not true of the decoder-only model due to the causal constraint and not true of the encoder-decoder and encoder-encoder models as both employ a source-side encoder that isolates and prevents it from conditioning on the target-side.  

One explanation for this behaviour is the encoder-only model may learn to write $t$ into the correct target position 
and then condition on this latent variable when forming the representation of $G$.  Such an algorithm would explain why 1-step NAR and $M$-step IAR inference both produce the same results, as the iterative process may already be applied in the latent space across layers.   This also relates to Exp.\@ 26x, which indicates that conditioning on ground-truths is not a critical component of such behaviour and that $t$ may be sufficient supervision given the appropriate parameterization.  

The path-star task is seemingly trivial but deceptively difficult and we make large headway into solving the mystery behind this difficulty.  While we do not solve it using the decoder-only model, we expose several issues that make the task harder but are not caused by teacher-forcing.  We demonstrate that it is solvable in the simplified setting using arm-wise permutation.   We show that the relative position of $Q$ to $G$ induces a representation problem for decoder-only models (Exp.\@ 4 vs.\@ 5). We explain why this works and why edge-wise permutation increases the difficulty via RASP. We also show that overfitting is an issue (Exp.\@ 4 vs.\@ 7).  

We show the task can be learnt, if inconsistently, via an AR encoder-decoder model trained using teacher-forcing (Exps.\@ 14 and 15).  These successes depend on using a non-causal representation of $G$ and structured samples.  Thus the difficulty of the task is reduced when accounting for these hidden factors.  The arm-wise exps.\@ show a key factor to the task's difficulty is the edge-wise permutation of $G$ and providing more structure to the graph makes the task solvable. 
Given this, we expect the issue with next-token prediction to apply to graph tasks where the graph needs to be reconstructed from a set of shuffled edges.  We are skeptical that it will apply to general tasks without these conditions.  

Lastly, we find the encoder-only model can consistently solve the task (Exp.\@ 24x). Thus the task is not pathological to teacher-forcing per se but specifically when combined with a causal AR model. 



\section{Limitations}


We do not claim to have fully unraveled the mystery of the path-star task and we try to fully exhibit the limitations of the current work, in hopes that others will become interested in improving on our results.    Solving the task is not important in itself, but rather it is the insights gained into why the task is difficult for certain models or training methods that is important. \citet{bachmann2024the} introduced the task because of the strong implications it would have to the next-token prediction paradigm if the task was not learnable due to the CHC.  By questioning the implications of the CHC hypothesis, we have potentially reduced the importance of the task.  This begs the question, is the path-star task still interesting?  One major limitation to our work is that we can not fully explain why the encoder-only model is the only one to work and why in both the NAR and IAR settings.  {\bf However, the fact that the four different types of models all have different performances and thus capabilities in regards to learning the task, shows that the path-star task is useful in exposing these different capabilities and learning dynamics, and thus justifies interest in the task.}    

One explanation is that the encoder-only model can use the target-side tokens for latent computation which can affect the source-side.  This is mentioned above but we do not attempt any experiments in this direction here.  
Given the $\mathcal{O}(\log M)$ RASP program requires the `|' tokens to be used as state-storage, we feel like extra compute tokens in the sequence may be required for efficiently learning the task.  This may be related to `thinking' tokens \citep{herel2024thinking, goyal2023think}.  
This is a direction we leave for future work. 

Arguably, the original task setting requires that we solve the problem using a decoder-only model.  We made progress in this, by considering the relative position of $Q$ to $G$ and proved decoder-only models can solve the task via RASP, but we have no empirical results where the solution is learnt. 

One missing experiment would be increasing the training dataset size.  We are implicitly doing this via structured samples which acts as a `smart' way of increasing the data. 
However, this method does not always prevent overfitting, so dataset size may still be an issue.  We do not consider large values of $D$.  We also only do 11 trials per experiment, which is not enough for significance.  However, this work does report the results of $31 \times 4 \times 11 = 1364$ trials.  

We only consider a single value of $M$.  This is justified in the fact that the RASP analysis indicates that the models will not generalize as $M$ increases with edge-wise permutation.  We do not believe that there exists an algorithm that can solve the task with $\mathcal{O}(1)$ layers, though we have not proven this. This brings us to an important open question: the IAR models do not find the reverse order solution. 
We expect this has to do with the uniform order-of-generation assumption they employ.  Interestingly, if they could find the reverse order, then it can be done with $\mathcal{O}(1)$ layers, which would show that IAR models can generalize to larger $M$ where AR and NAR models can not.  

Related to this is our focus on transformers.  The arm traversal requires $\mathcal{O}(\log M)$ or $\mathcal{O}(M)$ layers because transformers can only make pair-wise comparisons on tokens.  LSTMs have no such limitation and thus may be capable of encoding the arms better than transformers (but this may not be true \citep{pmlr-v235-sanford24a}).  However, LSTMs and transformers are known to behave differently to sensitive functions. 
Thus any potential differences between models will be interesting, but it may be difficult to pinpoint the reasons for the differences.

Finally, we attempt no experiments with LLMs.  If a LLM can solve the task, that is only interesting or informative insofar as it provides an explanation for why a large model can when a small one can not.  As such, we believe the results we have provided in this work will still be critical in such a case.  
If LLMs can solve the task, it will be because they are making use of some learnt bias which helps with the task, but given the opaque nature of LLMs and their training, this will be difficult to discern.

\section*{Acknowledgments}

We thank our supervisor 
Frank Rudzicz for financing going to EMNLP, Siavash Kazemian, Gagandeep Singh, Shunan Zhao, and Vaishnavh Nagarajan for their insightful feedback, Reviewer 3 who strongly advocated for our work, and Reviewer 1, who despite being apprehensive of our work, engaged in a long discussion with us, correctly pointed out issues, and provided strong feedback, allowing us to significantly improve our work.  


Resources used in preparing this research were provided, in part, by the Province of Ontario, the Government of Canada through CIFAR, and companies sponsoring the Vector Institute


\bibliography{main}

\begin{thebibliography}{41}
\providecommand{\natexlab}[1]{#1}

\bibitem[{Arora et~al.(2022)Arora, El~Asri, Bahuleyan, and Cheung}]{arora-etal-2022-exposure}
Kushal Arora, Layla El~Asri, Hareesh Bahuleyan, and Jackie Cheung. 2022.
\newblock \href {https://doi.org/10.18653/v1/2022.findings-acl.58} {Why exposure bias matters: An imitation learning perspective of error accumulation in language generation}.
\newblock In \emph{Findings of the Association for Computational Linguistics: ACL 2022}, pages 700--710, Dublin, Ireland. Association for Computational Linguistics.

\bibitem[{Bachmann and Nagarajan(2024)}]{bachmann2024the}
Gregor Bachmann and Vaishnavh Nagarajan. 2024.
\newblock \href {https://proceedings.mlr.press/v235/bachmann24a.html} {The pitfalls of next-token prediction}.
\newblock In \emph{Proceedings of the 41st International Conference on Machine Learning}, volume 235 of \emph{Proceedings of Machine Learning Research}, pages 2296--2318. PMLR.

\bibitem[{Bengio et~al.(2015)Bengio, Vinyals, Jaitly, and Shazeer}]{bengio2015scheduled}
Samy Bengio, Oriol Vinyals, Navdeep Jaitly, and Noam Shazeer. 2015.
\newblock Scheduled sampling for sequence prediction with recurrent neural networks.
\newblock In \emph{Proceedings of the 28th International Conference on Neural Information Processing Systems - Volume 1}, NIPS'15, page 1171–1179, Cambridge, MA, USA. MIT Press.

\bibitem[{Berglund et~al.(2024)Berglund, Tong, Kaufmann, Balesni, Stickland, Korbak, and Evans}]{berglund2024the}
Lukas Berglund, Meg Tong, Maximilian Kaufmann, Mikita Balesni, Asa~Cooper Stickland, Tomasz Korbak, and Owain Evans. 2024.
\newblock \href {https://openreview.net/forum?id=GPKTIktA0k} {The reversal curse: {LLM}s trained on {\textquotedblleft}a is b{\textquotedblright} fail to learn {\textquotedblleft}b is a{\textquotedblright}}.
\newblock In \emph{The Twelfth International Conference on Learning Representations}.

\bibitem[{Bhattamishra et~al.(2023)Bhattamishra, Patel, Kanade, and Blunsom}]{bhattamishra-etal-2023-simplicity}
Satwik Bhattamishra, Arkil Patel, Varun Kanade, and Phil Blunsom. 2023.
\newblock \href {https://doi.org/10.18653/v1/2023.acl-long.317} {Simplicity bias in transformers and their ability to learn sparse {B}oolean functions}.
\newblock In \emph{Proceedings of the 61st Annual Meeting of the Association for Computational Linguistics (Volume 1: Long Papers)}, pages 5767--5791, Toronto, Canada. Association for Computational Linguistics.

\bibitem[{Brown et~al.(2020)Brown, Mann, Ryder, Subbiah, Kaplan, Dhariwal, Neelakantan, Shyam, Sastry, Askell, Agarwal, Herbert-Voss, Krueger, Henighan, Child, Ramesh, Ziegler, Wu, Winter, Hesse, Chen, Sigler, Litwin, Gray, Chess, Clark, Berner, McCandlish, Radford, Sutskever, and Amodei}]{NEURIPS2020_1457c0d6}
Tom Brown, Benjamin Mann, Nick Ryder, Melanie Subbiah, Jared~D Kaplan, Prafulla Dhariwal, Arvind Neelakantan, Pranav Shyam, Girish Sastry, Amanda Askell, Sandhini Agarwal, Ariel Herbert-Voss, Gretchen Krueger, Tom Henighan, Rewon Child, Aditya Ramesh, Daniel Ziegler, Jeffrey Wu, Clemens Winter, Chris Hesse, Mark Chen, Eric Sigler, Mateusz Litwin, Scott Gray, Benjamin Chess, Jack Clark, Christopher Berner, Sam McCandlish, Alec Radford, Ilya Sutskever, and Dario Amodei. 2020.
\newblock \href {https://proceedings.neurips.cc/paper_files/paper/2020/file/1457c0d6bfcb4967418bfb8ac142f64a-Paper.pdf} {Language models are few-shot learners}.
\newblock In \emph{Advances in Neural Information Processing Systems}, volume~33, pages 1877--1901. Curran Associates, Inc.

\bibitem[{Bubeck et~al.(2023)Bubeck, Chandrasekaran, Eldan, Gehrke, Horvitz, Kamar, Lee, Lee, Li, Lundberg et~al.}]{bubeck2023sparks}
S{\'e}bastien Bubeck, Varun Chandrasekaran, Ronen Eldan, Johannes Gehrke, Eric Horvitz, Ece Kamar, Peter Lee, Yin~Tat Lee, Yuanzhi Li, Scott Lundberg, et~al. 2023.
\newblock Sparks of artificial general intelligence: Early experiments with gpt-4.
\newblock \emph{arXiv preprint arXiv:2303.12712}.

\bibitem[{Chakraborty et~al.(2023)Chakraborty, Kulkarni, and Li}]{chakraborty-etal-2023-zero}
Mohna Chakraborty, Adithya Kulkarni, and Qi~Li. 2023.
\newblock \href {https://doi.org/10.18653/v1/2023.acl-long.313} {Zero-shot approach to overcome perturbation sensitivity of prompts}.
\newblock In \emph{Proceedings of the 61st Annual Meeting of the Association for Computational Linguistics (Volume 1: Long Papers)}, pages 5698--5711, Toronto, Canada. Association for Computational Linguistics.

\bibitem[{Chen et~al.(2024)Chen, Chi, Wang, and Zhou}]{chen2024premise}
Xinyun Chen, Ryan~Andrew Chi, Xuezhi Wang, and Denny Zhou. 2024.
\newblock \href {https://openreview.net/forum?id=4zAHgkiCQg} {Premise order matters in reasoning with large language models}.
\newblock In \emph{Forty-first International Conference on Machine Learning}.

\bibitem[{Chen et~al.(2023)Chen, Zhao, Yu, McKeown, and He}]{chen-etal-2023-relation}
Yanda Chen, Chen Zhao, Zhou Yu, Kathleen McKeown, and He~He. 2023.
\newblock \href {https://doi.org/10.18653/v1/2023.findings-emnlp.12} {On the relation between sensitivity and accuracy in in-context learning}.
\newblock In \emph{Findings of the Association for Computational Linguistics: EMNLP 2023}, pages 155--167, Singapore. Association for Computational Linguistics.

\bibitem[{Devlin et~al.(2019)Devlin, Chang, Lee, and Toutanova}]{devlin-etal-2019-bert}
Jacob Devlin, Ming-Wei Chang, Kenton Lee, and Kristina Toutanova. 2019.
\newblock \href {https://doi.org/10.18653/v1/N19-1423} {{BERT}: Pre-training of deep bidirectional transformers for language understanding}.
\newblock In \emph{Proceedings of the 2019 Conference of the North {A}merican Chapter of the Association for Computational Linguistics: Human Language Technologies, Volume 1 (Long and Short Papers)}, pages 4171--4186, Minneapolis, Minnesota. Association for Computational Linguistics.

\bibitem[{Ghazvininejad et~al.(2019)Ghazvininejad, Levy, Liu, and Zettlemoyer}]{ghazvininejad-etal-2019-mask}
Marjan Ghazvininejad, Omer Levy, Yinhan Liu, and Luke Zettlemoyer. 2019.
\newblock \href {https://doi.org/10.18653/v1/D19-1633} {Mask-predict: Parallel decoding of conditional masked language models}.
\newblock In \emph{Proceedings of the 2019 Conference on Empirical Methods in Natural Language Processing and the 9th International Joint Conference on Natural Language Processing (EMNLP-IJCNLP)}, pages 6112--6121, Hong Kong, China. Association for Computational Linguistics.

\bibitem[{Golovneva et~al.(2024)Golovneva, Allen-Zhu, Weston, and Sukhbaatar}]{golovneva2024reverse}
Olga Golovneva, Zeyuan Allen-Zhu, Jason~E Weston, and Sainbayar Sukhbaatar. 2024.
\newblock \href {https://openreview.net/forum?id=HDkNbfLQgu} {Reverse training to nurse the reversal curse}.
\newblock In \emph{First Conference on Language Modeling}.

\bibitem[{Goyal et~al.(2024)Goyal, Ji, Rawat, Menon, Kumar, and Nagarajan}]{goyal2023think}
Sachin Goyal, Ziwei Ji, Ankit~Singh Rawat, Aditya~Krishna Menon, Sanjiv Kumar, and Vaishnavh Nagarajan. 2024.
\newblock Think before you speak: Training language models with pause tokens.
\newblock In \emph{The Twelfth International Conference on Learning Representations}.

\bibitem[{Gu and Dao(2023)}]{gu2023mamba}
Albert Gu and Tri Dao. 2023.
\newblock Mamba: Linear-time sequence modeling with selective state spaces.
\newblock \emph{arXiv preprint arXiv:2312.00752}.

\bibitem[{Gu et~al.(2018)Gu, Bradbury, Xiong, Li, and Socher}]{gu2018non}
Jiatao Gu, James Bradbury, Caiming Xiong, Victor~OK Li, and Richard Socher. 2018.
\newblock Non-autoregressive neural machine translation.
\newblock In \emph{International Conference on Learning Representations}.

\bibitem[{Gu and Kong(2021)}]{gu-kong-2021-fully}
Jiatao Gu and Xiang Kong. 2021.
\newblock \href {https://doi.org/10.18653/v1/2021.findings-acl.11} {Fully non-autoregressive neural machine translation: Tricks of the trade}.
\newblock In \emph{Findings of the Association for Computational Linguistics: ACL-IJCNLP 2021}, pages 120--133, Online. Association for Computational Linguistics.

\bibitem[{Hahn et~al.(2021)Hahn, Jurafsky, and Futrell}]{hahn2021sensitivity}
Michael Hahn, Dan Jurafsky, and Richard Futrell. 2021.
\newblock Sensitivity as a complexity measure for sequence classification tasks.
\newblock \emph{Transactions of the Association for Computational Linguistics}, 9:891--908.

\bibitem[{Hahn and Rofin(2024)}]{hahn2024sensitive}
Michael Hahn and Mark Rofin. 2024.
\newblock \href {https://doi.org/10.18653/v1/2024.acl-long.800} {Why are sensitive functions hard for transformers?}
\newblock In \emph{Proceedings of the 62nd Annual Meeting of the Association for Computational Linguistics (Volume 1: Long Papers)}, pages 14973--15008, Bangkok, Thailand. Association for Computational Linguistics.

\bibitem[{Haviv et~al.(2022)Haviv, Ram, Press, Izsak, and Levy}]{haviv-etal-2022-transformer}
Adi Haviv, Ori Ram, Ofir Press, Peter Izsak, and Omer Levy. 2022.
\newblock \href {https://doi.org/10.18653/v1/2022.findings-emnlp.99} {Transformer language models without positional encodings still learn positional information}.
\newblock In \emph{Findings of the Association for Computational Linguistics: EMNLP 2022}, pages 1382--1390, Abu Dhabi, United Arab Emirates. Association for Computational Linguistics.

\bibitem[{Herel and Mikolov(2023)}]{herel2024thinking}
David Herel and Tomas Mikolov. 2023.
\newblock Thinking tokens for language modeling.
\newblock \emph{8th Conference on Artificial Intelligence and Theorem Proving}.

\bibitem[{Kazemnejad et~al.(2023)Kazemnejad, Padhi, Natesan~Ramamurthy, Das, and Reddy}]{NEURIPS2023_4e85362c}
Amirhossein Kazemnejad, Inkit Padhi, Karthikeyan Natesan~Ramamurthy, Payel Das, and Siva Reddy. 2023.
\newblock \href {https://proceedings.neurips.cc/paper_files/paper/2023/file/4e85362c02172c0c6567ce593122d31c-Paper-Conference.pdf} {The impact of positional encoding on length generalization in transformers}.
\newblock In \emph{Advances in Neural Information Processing Systems}, volume~36, pages 24892--24928. Curran Associates, Inc.

\bibitem[{Lee et~al.(2018)Lee, Mansimov, and Cho}]{lee-etal-2018-deterministic}
Jason Lee, Elman Mansimov, and Kyunghyun Cho. 2018.
\newblock \href {https://doi.org/10.18653/v1/D18-1149} {Deterministic non-autoregressive neural sequence modeling by iterative refinement}.
\newblock In \emph{Proceedings of the 2018 Conference on Empirical Methods in Natural Language Processing}, pages 1173--1182, Brussels, Belgium. Association for Computational Linguistics.

\bibitem[{Liao et~al.(2020)Liao, Jiang, and Liu}]{liao-etal-2020-probabilistically}
Yi~Liao, Xin Jiang, and Qun Liu. 2020.
\newblock \href {https://doi.org/10.18653/v1/2020.acl-main.24} {Probabilistically masked language model capable of autoregressive generation in arbitrary word order}.
\newblock In \emph{Proceedings of the 58th Annual Meeting of the Association for Computational Linguistics}, pages 263--274, Online. Association for Computational Linguistics.

\bibitem[{Monea et~al.(2023)Monea, Joulin, and Grave}]{monea2023pass}
Giovanni Monea, Armand Joulin, and Edouard Grave. 2023.
\newblock Pass: Parallel speculative sampling.
\newblock \emph{arXiv preprint arXiv:2311.13581}.

\bibitem[{Nezhurina et~al.(2024)Nezhurina, Cipolina-Kun, Cherti, and Jitsev}]{nezhurina2024alice}
Marianna Nezhurina, Lucia Cipolina-Kun, Mehdi Cherti, and Jenia Jitsev. 2024.
\newblock Alice in wonderland: Simple tasks showing complete reasoning breakdown in state-of-the-art large language models.
\newblock \emph{arXiv preprint arXiv:2406.02061}.

\bibitem[{Ott et~al.(2019)Ott, Edunov, Baevski, Fan, Gross, Ng, Grangier, and Auli}]{ott2019fairseq}
Myle Ott, Sergey Edunov, Alexei Baevski, Angela Fan, Sam Gross, Nathan Ng, David Grangier, and Michael Auli. 2019.
\newblock fairseq: A fast, extensible toolkit for sequence modeling.
\newblock \emph{arXiv preprint arXiv:1904.01038}.

\bibitem[{Qian et~al.(2021)Qian, Zhou, Bao, Wang, Qiu, Zhang, Yu, and Li}]{qian-etal-2021-glancing}
Lihua Qian, Hao Zhou, Yu~Bao, Mingxuan Wang, Lin Qiu, Weinan Zhang, Yong Yu, and Lei Li. 2021.
\newblock \href {https://doi.org/10.18653/v1/2021.acl-long.155} {Glancing transformer for non-autoregressive neural machine translation}.
\newblock In \emph{Proceedings of the 59th Annual Meeting of the Association for Computational Linguistics and the 11th International Joint Conference on Natural Language Processing (Volume 1: Long Papers)}, pages 1993--2003, Online. Association for Computational Linguistics.

\bibitem[{Radford et~al.()Radford, Wu, Child, Luan, Amodei, Sutskever et~al.}]{radford2019language}
Alec Radford, Jeffrey Wu, Rewon Child, David Luan, Dario Amodei, Ilya Sutskever, et~al.
\newblock Language models are unsupervised multitask learners.

\bibitem[{Ranzato et~al.(2016)Ranzato, Chopra, Auli, and Zaremba}]{ranzato2016sequence}
Marc’Aurelio Ranzato, Sumit Chopra, Michael Auli, and Wojciech Zaremba. 2016.
\newblock Sequence level training with recurrent neural networks.
\newblock In \emph{4th International Conference on Learning Representations, ICLR 2016}.

\bibitem[{Sanford et~al.(2024)Sanford, Hsu, and Telgarsky}]{pmlr-v235-sanford24a}
Clayton Sanford, Daniel Hsu, and Matus Telgarsky. 2024.
\newblock \href {https://proceedings.mlr.press/v235/sanford24a.html} {Transformers, parallel computation, and logarithmic depth}.
\newblock In \emph{Proceedings of the 41st International Conference on Machine Learning}, volume 235 of \emph{Proceedings of Machine Learning Research}, pages 43276--43327. PMLR.

\bibitem[{Tsai et~al.(2019)Tsai, Bai, Yamada, Morency, and Salakhutdinov}]{tsai-etal-2019-transformer}
Yao-Hung~Hubert Tsai, Shaojie Bai, Makoto Yamada, Louis-Philippe Morency, and Ruslan Salakhutdinov. 2019.
\newblock \href {https://doi.org/10.18653/v1/D19-1443} {Transformer dissection: An unified understanding for transformer{'}s attention via the lens of kernel}.
\newblock In \emph{Proceedings of the 2019 Conference on Empirical Methods in Natural Language Processing and the 9th International Joint Conference on Natural Language Processing (EMNLP-IJCNLP)}, pages 4344--4353, Hong Kong, China. Association for Computational Linguistics.

\bibitem[{Valmeekam et~al.(2023)Valmeekam, Marquez, Sreedharan, and Kambhampati}]{valmeekam2023on}
Karthik Valmeekam, Matthew Marquez, Sarath Sreedharan, and Subbarao Kambhampati. 2023.
\newblock \href {https://openreview.net/forum?id=X6dEqXIsEW} {On the planning abilities of large language models - a critical investigation}.
\newblock In \emph{Thirty-seventh Conference on Neural Information Processing Systems}.

\bibitem[{Vaswani et~al.(2017)Vaswani, Shazeer, Parmar, Uszkoreit, Jones, Gomez, Kaiser, and Polosukhin}]{vaswani2017attention}
Ashish Vaswani, Noam Shazeer, Niki Parmar, Jakob Uszkoreit, Llion Jones, Aidan~N Gomez, {\L}ukasz Kaiser, and Illia Polosukhin. 2017.
\newblock Attention is all you need.
\newblock \emph{Advances in neural information processing systems}, 30.

\bibitem[{Wang et~al.(2018)Wang, Zhang, and Chen}]{wang-etal-2018-semi-autoregressive}
Chunqi Wang, Ji~Zhang, and Haiqing Chen. 2018.
\newblock \href {https://doi.org/10.18653/v1/D18-1044} {Semi-autoregressive neural machine translation}.
\newblock In \emph{Proceedings of the 2018 Conference on Empirical Methods in Natural Language Processing}, pages 479--488, Brussels, Belgium. Association for Computational Linguistics.

\bibitem[{Weiss et~al.(2021)Weiss, Goldberg, and Yahav}]{weiss2021thinking}
Gail Weiss, Yoav Goldberg, and Eran Yahav. 2021.
\newblock Thinking like transformers.
\newblock In \emph{International Conference on Machine Learning}, pages 11080--11090. PMLR.

\bibitem[{Wies et~al.(2023)Wies, Levine, and Shashua}]{wies2023subtask}
Noam Wies, Yoav Levine, and Amnon Shashua. 2023.
\newblock \href {https://openreview.net/forum?id=BrJATVZDWEH} {Sub-task decomposition enables learning in sequence to sequence tasks}.
\newblock In \emph{The Eleventh International Conference on Learning Representations}.

\bibitem[{Williams and Zipser(1989)}]{williams1989learning}
Ronald~J Williams and David Zipser. 1989.
\newblock A learning algorithm for continually running fully recurrent neural networks.
\newblock \emph{Neural computation}, 1(2):270--280.

\bibitem[{Yang et~al.(2019)Yang, Dai, Yang, Carbonell, Salakhutdinov, and Le}]{yang2019xlnet}
Zhilin Yang, Zihang Dai, Yiming Yang, Jaime Carbonell, Russ~R Salakhutdinov, and Quoc~V Le. 2019.
\newblock Xlnet: Generalized autoregressive pretraining for language understanding.
\newblock \emph{Advances in neural information processing systems}, 32.

\bibitem[{Zhou et~al.(2024)Zhou, Bradley, Littwin, Razin, Saremi, Susskind, Bengio, and Nakkiran}]{zhou2024what}
Hattie Zhou, Arwen Bradley, Etai Littwin, Noam Razin, Omid Saremi, Joshua~M. Susskind, Samy Bengio, and Preetum Nakkiran. 2024.
\newblock \href {https://openreview.net/forum?id=AssIuHnmHX} {What algorithms can transformers learn? a study in length generalization}.
\newblock In \emph{The Twelfth International Conference on Learning Representations}.

\bibitem[{Zoph et~al.(2022)Zoph, Raffel, Schuurmans, Yogatama, Zhou, Metzler, Chi, Wei, Dean, Fedus, Bosma, Vinyals, Liang, Borgeaud, Hashimoto, and Tay}]{52065}
Barret Zoph, Colin Raffel, Dale Schuurmans, Dani Yogatama, Denny Zhou, Don Metzler, Ed~H. Chi, Jason Wei, Jeff Dean, Liam~B. Fedus, Maarten~Paul Bosma, Oriol Vinyals, Percy Liang, Sebastian Borgeaud, Tatsunori~B. Hashimoto, and Yi~Tay. 2022.
\newblock Emergent abilities of large language models.
\newblock \emph{TMLR}.

\end{thebibliography}

\appendix



\section{Model Details}

\begin{figure}[!htbp]
    \centering
    \includegraphics[width=1.\linewidth]{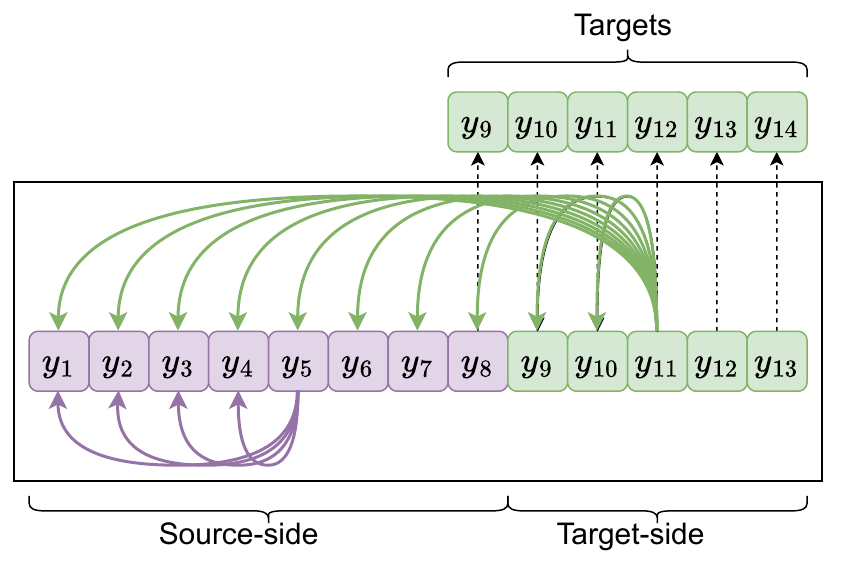}
    \caption{A decoder-only model.}
    \label{fig:decoder-only-model}
\end{figure}

\begin{figure}[!htbp]
    \centering
    \includegraphics[width=1.\linewidth]{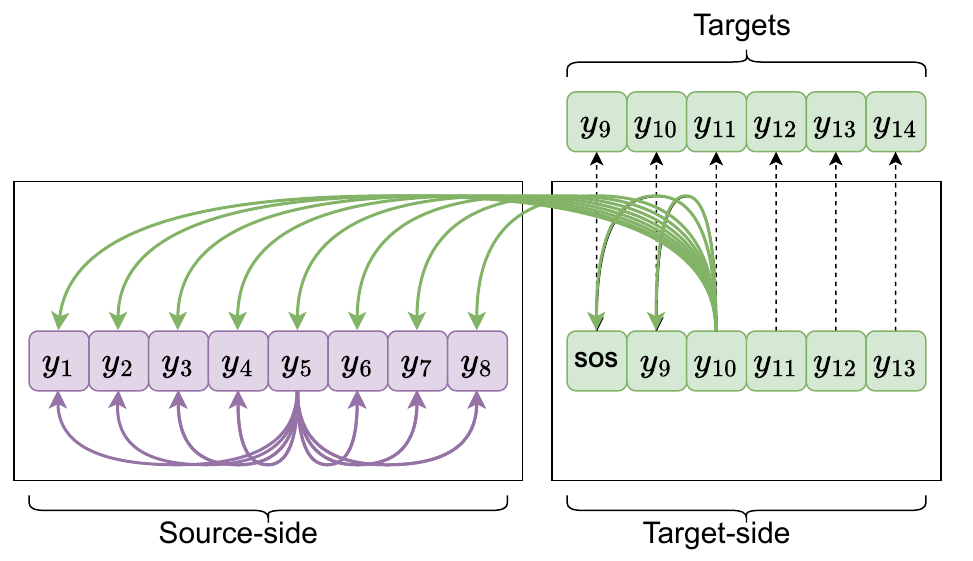}
    \caption{An encoder-decoder model.}
    \label{fig:encoder-decoder-model}
\end{figure}

\begin{figure}[!htbp]
    \centering
    \includegraphics[width=1.\linewidth]{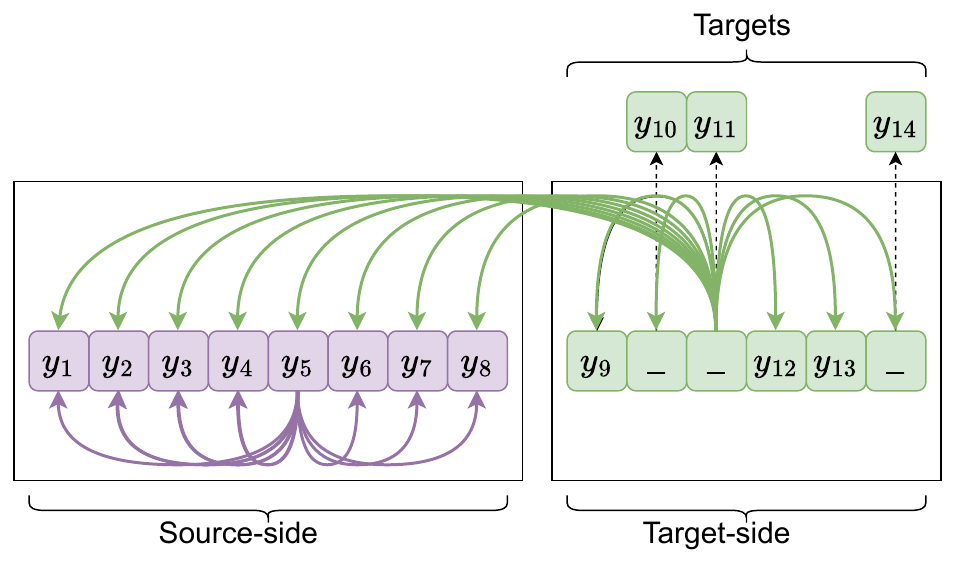}
    \caption{An encoder-encoder IAR model.}
    \label{fig:encoder-encoder-model}
\end{figure}

\begin{figure}[!htbp]
    \centering
    \includegraphics[width=1.\linewidth]{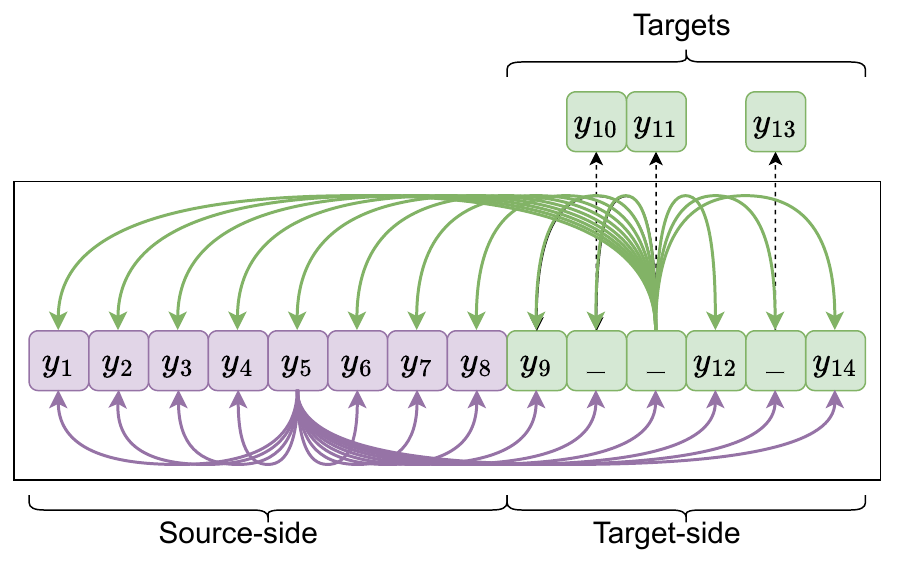}
    \caption{An encoder-only IAR model.}
    \label{fig:encoder-only-model}
\end{figure}

In Figures \ref{fig:decoder-only-model}, \ref{fig:encoder-decoder-model}, \ref{fig:encoder-encoder-model}, \ref{fig:encoder-only-model} we show the different parameterizations and attention constraints between models.  We illustrate all possible attention connections for one token on the source-side (lilac) and one token on the target-side (green).  For the encoder-encoder model, we input the partial ground-truth $y_9,\, y_{12},\,  y_{13}$ with $y_{10},\, y_{11},\, y_{14}$ masked and for the encoder-only model we input the partial ground-truth $y_9,\, y_{12},\, y_{14}$ with $y_{10},\, y_{11},\, y_{13}$ masked.  These represent different masking samples during training.  Note that the prediction of $y_{11}$ in both cases does not depend on the ground-truth of $y_{10}$ and hence bypases the CHC.

\subsection{Inference details}\label{app:inference}

Traditional AR models employ an autoregressive inference procedure where the predicted output is used as part of the conditioning information for the proceeding step.  We use a `teacher-forced' inference procedure that conditions on the partial ground-truths instead.  We call it `teacher-forced' as this conditioning matches how teacher-forcing is done during training.  This means we are evaluating the ground-truth-conditioned performance and not autoregressive performance during inference.  This evaluation method was also used by \citet{bachmann2024the} to prevent or rule out any inference-time bias.  We also performed traditional autoregressive inference, however, this produced the same results {\em in terms of sequence accuracy} (and thus we did not report them).  This is because as soon as the wrong leading node is predicted all succeeding nodes are incorrectly predicted and thus, the sequence accuracy is the same as $l_t$ is incorrect in both cases. Thus the reason we can do this is because we end up with the same results as traditional autoregressive inference and by doing it, we can explicitly rule out any issues stemming from training and inference time differences. 

Note this only applies to AR models.  The NAR models have no autoregressive procedure (and hence can not have an inference-time bias).  As stated above, we use 1-step NAR inference (with IAR training) for the IAR models.  Also, as mentioned, we tried true autoregressive $M$-step iterative inference for them as well, however, we did not report these results as they are the same as the 1-step NAR inference results.  Note we can not have a `teacher-forced inference' for the IAR models since there is no canonical order to partially condition and so we would have to sample different conditioning ground-truths which would not be an informative evaluation method.

\section{Task Representation and Tokenization}\label{appx:data}

\begin{figure*}[!htbp]
    \centering
    \includegraphics[width=1.\linewidth]{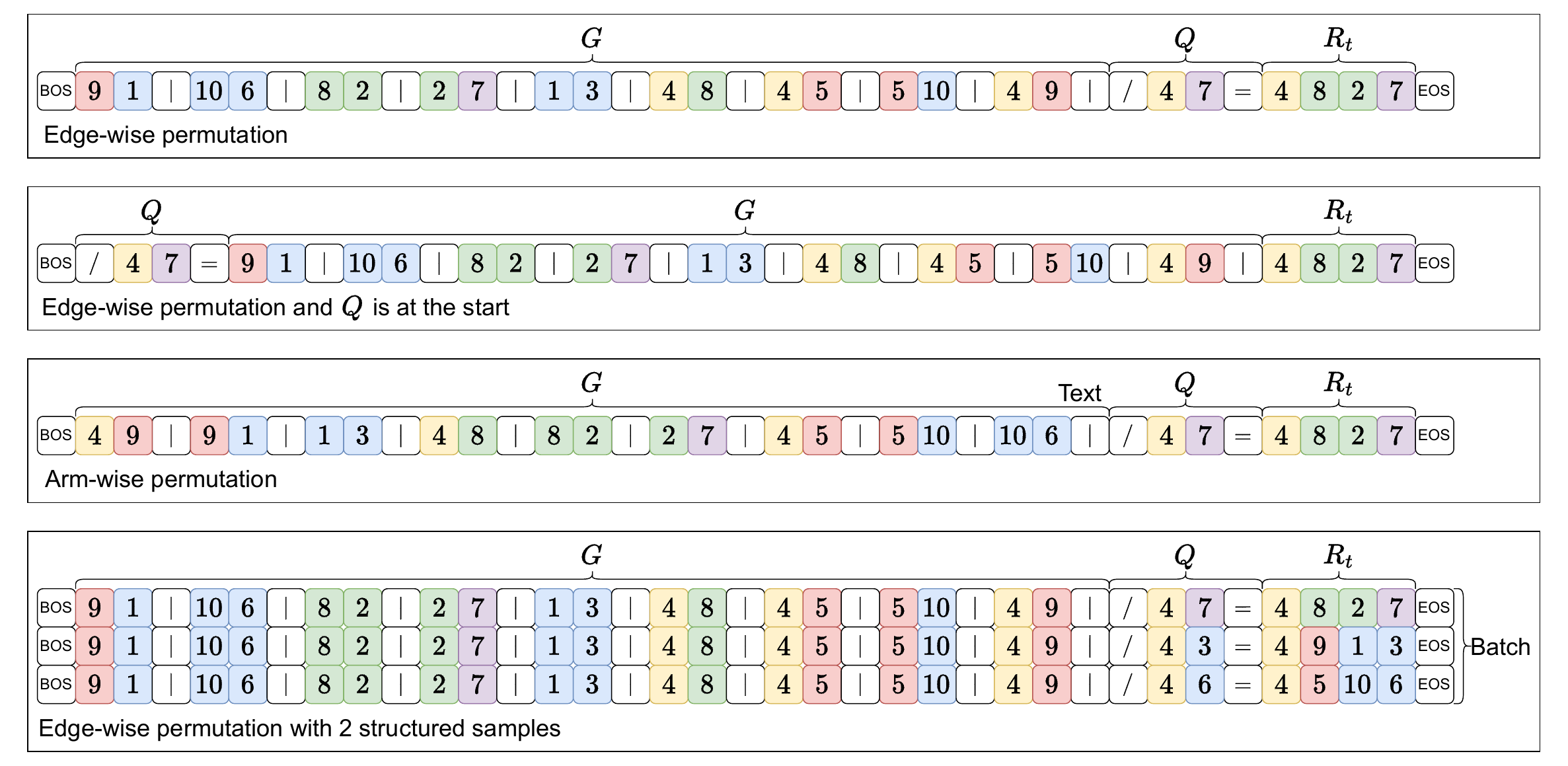}
    \caption{Different tokenizations for a given path-star graph.}
    \label{fig:data}
\end{figure*}

In Figure \ref{fig:data}, we illustrate multiple possible tokenizations in reference to the example path-star graph in Fig.\@ \ref{fig:psg}.  The first and third tokenizations are the same edge- and arm-wise examples provided in the caption of Fig.\@ \ref{fig:psg}.   The second one shows $Q$ before $G$, i.e.\@. `Start'.  The fourth tokenization shows (the only) two possible structured samples of the original edge-wise tokenization.

\section{The Clever Hans Phenomenon}
\label{app:CHC}

\begin{figure*}[!htbp]
    \centering
    \includegraphics[width=1.\linewidth]{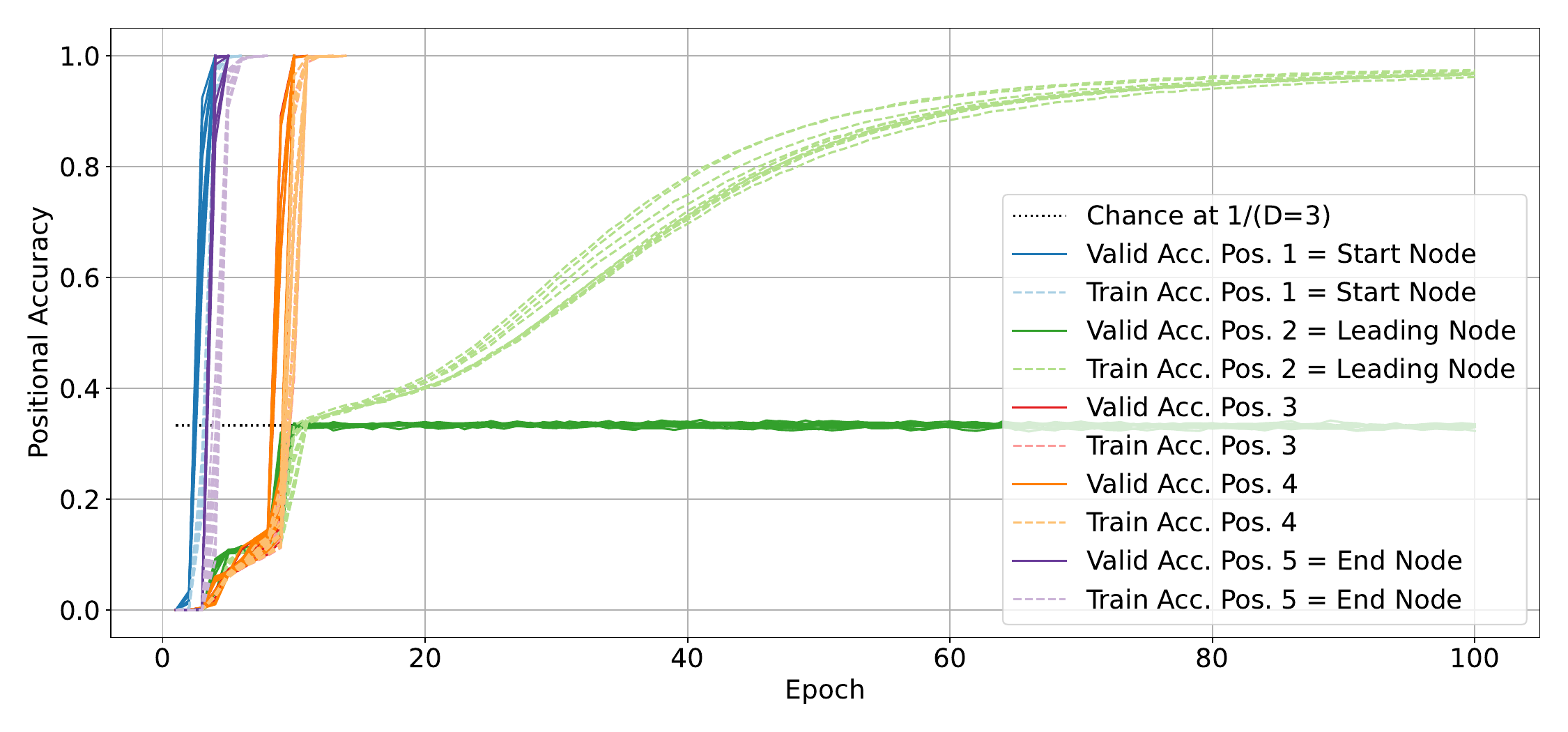}
    \caption{The appearance of the Clever Hans cheat over training. Data corresponds to row/Exp.\@ 1 of Table \ref{tab:decoder-only}, where $D$=3, $M=5$.}
    \label{fig:CHC}
\end{figure*}

In Fig \ref{fig:CHC} we show how the CHC appears during training.   Here we see that the first token to fit to 100\% accuracy is the given start node, $s$.  The next token is the given target node, $t$.  While this might seem strange as it is generated at the end of the sequence, this token is actually easily predictable since the model can infer that the target token should always be placed in the $M^{\text{th}}$ position.  This is because there is no requirement that predictions be generalizable to different arm lengths and hence the target token is always in the $M^{\text{th}}$ position.  This is explicitly done to maintain that the test data is in-domain with the training data.  Next, we see that all other non-leading nodes fit via the CHC.  As no trials succeeded in this experiment, the validation accuracy of the leading token becomes stagnant at chance, while the training accuracy improves (due to overfitting).  

\section{Structured Samples and Overfitting}\label{appx:samples}

\begin{figure*}[!htbp]
    \centering
    \includegraphics[width=1.\linewidth]{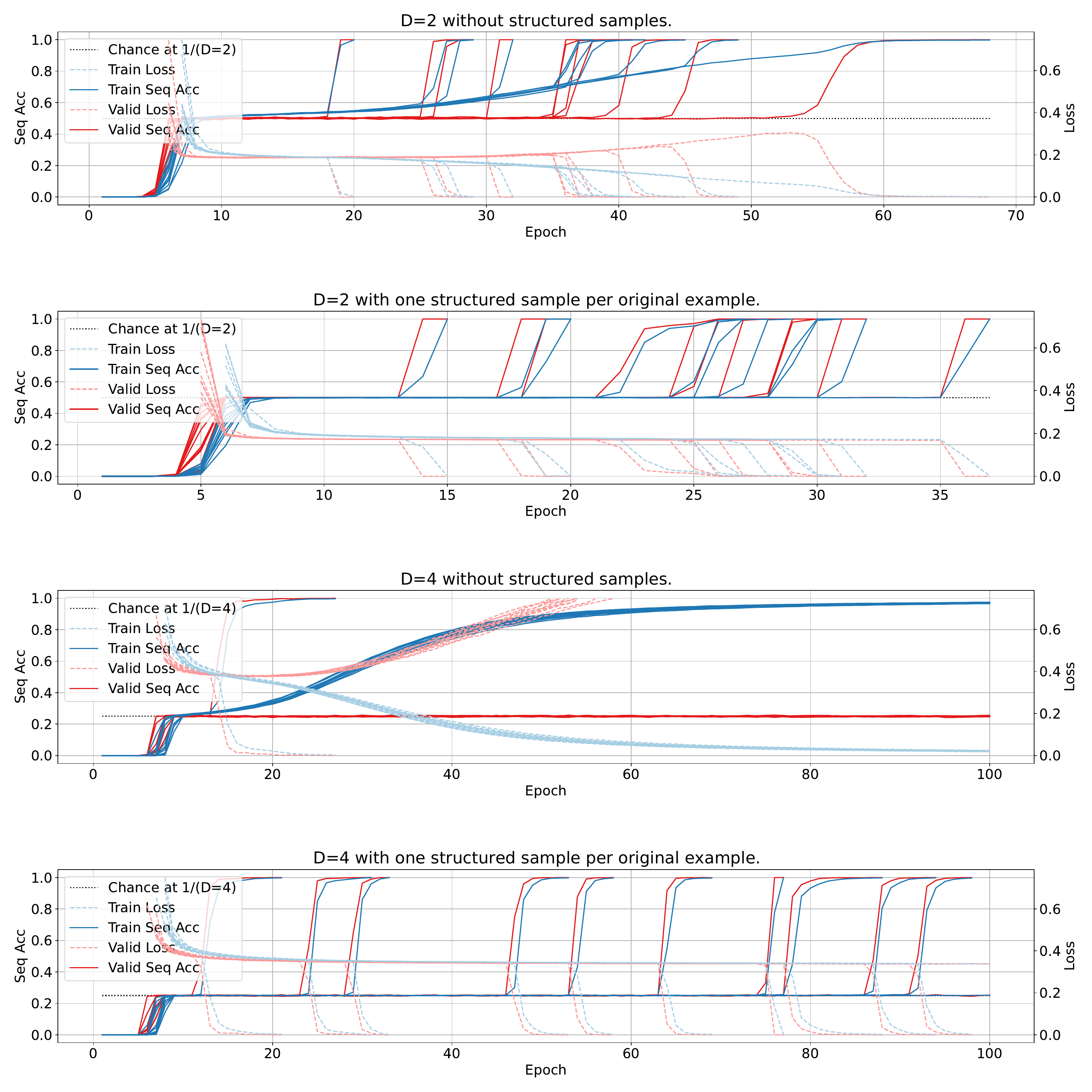}
    \caption[Overfitting and structured samples]{Plots 1 and 3 visualize the training of the experiments of Exp.\@ 4 in Table \ref{tab:decoder-only} where $D=2$ and $D=4$ respectively. Plots 2 and 4 visualize the corresponding experiments in Exp.\@ 7  when structured samples are employed.  Each plot shows the loss and sequence accuracy across all 11 trials of the given experiment for both the training and validation partitions.  When a trial succeeds in finding the desired solution, the sequence accuracy spikes to 100\%, and the validation loss drops to near-zero. The loss is cut off at 0.75 for visibility. 
    
     \hspace{\parindent} In plot 1, all trials succeed, however, when $D$ is increased to 4, only 1/11 trials succeed as shown in plot 3.  Here we see that the training and validation losses diverge shortly after epoch 20, resulting in overfitting.  In Plot 4, the use of structured samples prevents this divergence, leading to 10/11 trials succeeding, with the remaining trial not finding the solution within the 100 epoch limit.        
    }
    \label{fig:contra}
\end{figure*}

Fig \ref{fig:contra} shows the effect structured samples have on overfitting.  Note, that the number of structured samples must be less than or equal to $D-1$, as there are only $D$ target-arm pairs and we discount the original sample.

\section{RASP}\label{appx:RASP}

RASP (Restricted Access Sequence Programming) is a formal programming language used to validate the existence of a transformer that can solve a given task.   Our RASP code is based on the numpy-like version of RASP provided by \citet{zhou2024what}.  They extended RASP to causal attention.  They also place extra limitations on operations to disallow difficult-to-learn representations.  One such limitation restricts math on positional indices to be only single increments.  Our algorithms do not meet this limitation, however, we do not believe it causes issues with learnability for our specific task.   RASP programs take in a tokenized word/symbol sequence as input called `seq' and can create the indices, `idx', via Listing \ref{func:inidices}.   We always assume that $Q$ (`q' in RASP code) is at the start (this only matters for the causal algorithm in Listing 
\ref{func:causal-propagate}.      

`d' is a symbol-to-id dictionary  and `reverse\_d' is the inverted dictionary.  In some algorithms, we use extra special symbols to help present the output of the algorithm, however, these do not contribute to the solution and are only used for human readability. These include a junk symbol `j' and arm markers `a0' through `a$M-1$', which mark a node as being the $i^\text{th}$ place in the arm.  We use `-99' and `-89' as masking values.  Instead of using numeric ids for $s$ and $t$, we use the characters `s' and `t' for readability.      

Note, we only solve the task for identifying the leading node, where we say the RASP program solves the task if it can transform the input sequence such that the new sequence contains at least one `edge' or sequence of three contiguous tokens, $(i, j, k)$, where one of those tokens is the correct leading node, $l_t$, and the other is the indicated target node, $t$.   The algorithm for identifying the leading node is different from all other nodes, which can be identified using the CHC.  As such, even more layers may be required for the transformer since some may be allocated to leaning the CHC (potentially both can be learnt by the same layers but using different attention heads).  We do not claim that these are the optimal algorithms, 
as we just want to show that various solutions exist. 


In Appx.\@ \ref{appx:RASP-path-star} we outline algorithms for solving the path-star task and core and library functions (slightly modified from from \citet{zhou2024what}) are in Appx. \ref{appx:rasp-lib}.







\subsection{Path-star RASP programs}\label{appx:RASP-path-star}

RASP programs in Listings \ref{func:non-causal-propagate-back-target}, \ref{func:non-causal-propagate-backward-targets}, \ref{func:non-causal-propagate-forward-start}, and
\ref{func:non-causal-propagate-log} are for non-causal models (encoder-decoder\footnote{we count this as non-causal here as the computations are done on the encoder-side.}, encoder-encoder, or encoder-only).  They assume that $Q$ is before $G$, but only for convenience.  The first three require $\mathcal{O}(M)$ KQV operations and thus layers to route information across the arms. The program in Listing \ref{func:non-causal-propagate-log} requires $\mathcal{O}(\log M)$ layers. 

The RASP program in Listing \ref{func:non-causal-propagate-back-target} identifies the correct target edge and walks back the target node to the matching leading node.  This makes use of the BOS token to store the current state of the walk-back.
We use some helper values and states to make processing the output of each KQV operation easier.
`masked\_x' will mask out $Q$ from the input sequence and `g\_start' is the token offset due to $Q$ being at the start. 
`i\_nodes' and `j\_nodes' are further masked versions of the input such that only `i' and `j' position nodes are visible in the sequence respectively. 
This means that one can match a particular j-node with a particular i-node by value (or reverse) uniquely for all nodes except the start node.  We only show how these are computed for Listing \ref{func:non-causal-propagate-back-target} and just give them as function input for the others which also use them.  The RASP program in Listing \ref{func:non-causal-propagate-backward-targets} is similar, except all final nodes are walked back in parallel, and the one in \ref{func:non-causal-propagate-forward-start} does this but in the opposite direction, by identifying the leading nodes and walking them to the final nodes of each arm.   Since there is only one BOS token, we can not use it to keep the current state of the $D$ arms.  Instead, we overwrite the values of either the $i$ or $j$ nodes.  This is permitted as any edge is no longer useful for solving the problem once it has been traversed.     

The RASP program in Listing \ref{func:non-causal-arms-propagate-backward-targets} requires a constant, $\mathcal{O}(1)$, number of operations but requires that arm-wise permutation is used and does not work with edge-wise permutation.  This algorithm can just identify the final nodes and then look back at the appropriate number of positions via learning `leading\_offset' to find the correct leading node.

The RASP program in Listing \ref{func:non-causal-propagate-log} requires $\mathcal{O}(\log M)$ layers.  It makes use of two `|' markers instead of one, allowing us to traverse the edges in both the forward and backward directions.  These extra states are used to store the current current distance a given edge has been able to be routed to at each step.  We can not use the original $(i, j)$ node to store the arm states since, as we {\em skip connections}, we sometimes need the original edges to potentially do an extra step to match the leading nodes to the final nodes.  This depends on the length of the arms.  Note, using two `|' markers is not necessary but we believe using both directions makes the algorithm's behaviour more obvious.  Also, multiple heads would allow both directions to be considered by a single KQV operation.  

The logarithmic behaviour of this algorithm is achieved by {\em skipping} or {\em doubling} the routed distance at each step.   Since all connections are computed in parallel, the first step computes all direct connections.  The next step can then reuse these computations to compute the nodes 2-connections away from any given node.  And then this can be used to compute the nodes 4-connections away and so forth.  Note that if, at any step, the connections run out for a given position, this process stops, hence $M$ does not need to be an exact power of 2.  This parallel doubling algorithm was also found by \citet{pmlr-v235-sanford24a} to solve the $k$-hops task in $\mathcal{O}(\log k)$.

The RASP program in Listing \ref{func:causal-propagate} works for causal models with edge-wise permutation.  It requires that $Q$ is before $G$ and also requires $\mathcal{O}(M)$ operations.   This algorithm is similar to the ones in Listings \ref{func:non-causal-propagate-backward-targets} and \ref{func:non-causal-propagate-forward-start}, however, it is more complicated
due to the casual constraint.  The issue here is that we can only move towards the end the of the sequence.  Here we use the `|' tokens to keep the current leading nodes.   To achieve this we need to consider two different cases or rules.  In rule 1 we have the condition that the connecting edge is before the current edge. In this case, we move the next connecting edge's next token to the current edge.  In rule 2 we have the condition that the connecting edge is after the current one.  In that case, we copy the leading node over to this edge and mark it as the current edge.      

Note, that one explanation as to why $Q$ is needed to be before $G$ is that this allows us to identify the leading nodes via the start node.  One may have thought that providing the start node is redundant in $Q$, however, otherwise, the start node is only identifiable by determining that it is the only node with degree $D$.  However, the causal constraint means we can not count the number of instances of a token at all time steps and thus not its degree.  This would explain why placing $Q$ before $G$ makes the task easier for the decoder-only model.

\begin{figure*}[t]
\begin{lstlisting}[language=Python, label=func:non-causal-propagate-back-target, caption=Simple RASP program for non-causal encoders which propagates the given target node back across the arm until the leading node is found. ]

def non_causal_propagate_back_target(seq):
    # helper representations which can be deduced via the positions
    idx = indices(seq)
    masked_x = specification_mask(seq)  # mask out q
    
    # each node id will be unique within i_nodes and j_nodes
    # this can be deduced via positional information
    # keeping these as separate states just means that 
    # we do not need to do the masking at each step to process the kqv
    
    g_start = len(q) if q_before_g else 0
    i_nodes = where((idx - g_start) % 3 == 0, masked_x, full(masked_x, d['j']))
    j_nodes = where((idx - g_start) % 3 == 1, masked_x, full(masked_x, d['j']))
    
    write_in_mask = full(seq, False)
    write_in_mask[0] = True
    
    # look up the target token via position in specification first
    target_idx = kqv(j_nodes, full(j_nodes, d['t']), idx, equals, 
                     default=-99, causal=False)  # index in g
    cur_state = where(write_in_mask, target_idx, seq)
    
    for step in range(0, arm_len - 2):  # skip start and initial node
        cur_idx = where(write_in_mask, cur_state, full(cur_state, -99))
    
        # write in the arm token, this is just to show how an arm could be marked
        cur_state = where(cur_idx[0] == idx, 
                          full(cur_state, d[get_cur_marker(step)]), cur_state)
    
        # Get the connecting edge token
        cur_idx = where(write_in_mask, cur_state, full(cur_state, -100)) - 1
        cur_state = where(cur_idx[0] == idx, 
                          full(cur_state, d[get_cur_marker(step, True)]), cur_state)
    
        connecting_token = kqv(idx, cur_idx, i_nodes, equals, 
                               default=-99, causal=False)
        cur_state = where(write_in_mask, connecting_token, cur_state)
    
        # get connecting node idx
        if step < arm_len - 3:
            cur_idx = kqv(j_nodes, connecting_token, idx, equals, 
                          default=-99, causal=False)
            cur_state = where(write_in_mask, cur_idx, cur_state)
    
    assert reverse_d[cur_state[0]] == correct_initial_node
    return cur_state
\end{lstlisting}
\end{figure*}

\begin{figure*}[t]
\begin{lstlisting}[language=Python, label=func:non-causal-propagate-backward-targets, caption=RASP program for non-causal encoders which propogates each final node to the edge containing the leading node of each arm. ]

def non_causal_propagate_backward_targets(seq, idx, masked_x, 
                                          g_start, i_nodes, j_nodes):

    # is final if only once in masked_x
    counts = sel_width(select(masked_x, masked_x, equals, causal=False))
    is_final = equals(counts, full(counts, 1))

    # pos of final_nodes nodes in i_nodes
    final_nodes = where(is_final, j_nodes, full(idx, -99)) 
    final_idx = where(is_final, idx, full(idx, -99))
    final_idx_slash = where(is_final, idx + 1, full(idx, -99))  
    is_final_slash = kqv(idx + 1, idx, is_final, equals, 
                         default=False, causal=False)  # shift right
                         
    cur_state = seq
    connecting_nodes = kqv(final_idx_slash, idx, final_nodes, equals, 
                           default=-99, causal=False)  # shift
    for step in range(0, arm_len - 2):  # while start not found
        # find connecting indices
        connecting_idxs = kqv(j_nodes, connecting_nodes, idx, equals, 
                              default=-99, causal=False)

        cur_state = where(is_final_slash, connecting_idxs, seq)
        connecting_nodes = kqv(idx, connecting_idxs - 1, i_nodes, equals, 
                               default=-99, causal=False)
        cur_state = where(is_final_slash, connecting_nodes, seq)

    # check valid, assume q before g
    output = detokenize(cur_state)
    edges = [(output[i], output[i+1], output[i+2]) 
             for i in range(g_start, len(output), 3)]
    is_valid = False
    for edge in edges:
        if edge[1] == 't' and edge[2] == correct_initial_node:
            is_valid = True
            break
    assert is_valid
    return cur_state
\end{lstlisting}
\end{figure*}

\begin{figure*}[t]
\begin{lstlisting}[language=Python, label=func:non-causal-propagate-forward-start, caption=RASP program for non-causal encoders which propogates each leading node to the edge containing end node of each arm. ]

def non_causal_propagate_forward_start(seq, idx, masked_x, 
                                       g_start, i_nodes, j_nodes):
    # get start indices
    is_start = equals(i_nodes, full(j_nodes, .d['s']))
    # pos of start nodes in i_nodes
    start_idx = where(is_start, idx, full(idx, -99))
    cur_state = where(is_start, start_idx, .seq)

    # get leading nodes from j_nodes
    leading_nodes = kqv(idx, start_idx + 1, j_nodes, equals, 
                        default=-99, causal=False)
    cur_state = where(is_start, leading_nodes, seq)

    connecting_nodes = leading_nodes
    for step in range(0, arm_len - 2):  # while last not found
        # find connecting indices
        connecting_idxs = kqv(i_nodes, connecting_nodes, idx, equals, 
                              default=-99, causal=False)
        cur_state = where(is_start, connecting_idxs, seq)
        connecting_nodes = kqv(idx, connecting_idxs + 1, j_nodes, equals, 
                               default=-99, causal=False)
        cur_state = where(is_start, connecting_nodes, seq)

    # check valid, assume q before g
    output = detokenize(cur_state)
    edges = [(output[i], output[i+1]) for i in range(g_start, len(output), 3)]
    is_valid = False
    for edge in edges:
        if edge[0] == 't' and edge[1] == correct_initial_node:
            is_valid = True
            break
    assert is_valid
    return cur_state
\end{lstlisting}
\end{figure*}

\begin{figure*}[t]
\begin{lstlisting}[language=Python, label=func:non-causal-propagate-log, caption=RASP program for non-causal encoders with $\mathcal{O}(\log M)$ required layers.]
    def non_causal_propagate_log(seq, idx, masked_x, 
                                 g_start, i_nodes, j_nodes):

        # each edge is (i, j, k1, k2) due to having two '/' tokens
        # initialize each edge by moving i -> k1, j-> k2,
        i_nodes_pos = where(not_equals(i_nodes, full(idx,  d['j'])), 
                            idx, full(idx, -99))
        j_nodes_pos = where(is_true(j_nodes, full(idx,  d['j'])), 
                            idx, full(idx, -99))
        k1_nodes_pos = kqv(idx, i_nodes_pos + 2, idx, equals, 
                           default=-89, causal=False)
        k2_nodes_pos = kqv(idx, j_nodes_pos + 2, idx, equals, 
                           default=-99, causal=False)
        k1_nodes = kqv(k1_nodes_pos, idx, i_nodes, equals, 
                       default=-89, causal=False)
        k2_nodes = kqv(k2_nodes_pos, idx, j_nodes, equals, 
                       default=-99, causal=False)

        for step in range(0, int(np.ceil(np.log2(arm_len - 1)))):
            # if this k1 == other k2, other k1 -> this k1
            connecting_k1_pos = kqv(k2_nodes, k1_nodes, idx - 1, equals, 
                                    default=-89, causal=False)
            new_k1_nodes = kqv(idx, connecting_k1_pos, k1_nodes, equals, 
                               default=-89, causal=False)
            k1_nodes = where(is_true(new_k1_nodes), new_k1_nodes, k1_nodes)

            # if this k2 == other k1, other k2 ->  this k2
            connecting_k2_pos = kqv(k1_nodes, k2_nodes, idx + 1, equals, 
                                    default=-99, causal=False)
            new_k2_nodes = kqv(idx, connecting_k2_pos, k2_nodes, equals, 
                               default=-99, causal=False)
            # note we can do k1_nodes here if we want parallel processing
            k2_nodes = where(is_true(new_k2_nodes), new_k2_nodes, k2_nodes)

            # write in state
            cur_state = where(is_true(k1_nodes), k1_nodes, seq)
            cur_state = where(is_true(k2_nodes), k2_nodes, cur_state)

        # potentially do one extra step due to step size issues
        # if j = k1, then k2 -> j
        conn = kqv(k1_nodes, j_nodes, idx + 1, equals, 
                   default=-99, causal=False)
        new_j_nodes = kqv(idx, conn, k2_nodes, equals, 
                          default=-99, causal=False)
        j_nodes = where(is_true(new_j_nodes), new_j_nodes, j_nodes)
        cur_state = where(is_true(j_nodes_pos), j_nodes, cur_state)

        # check valid, assume q before g
        # note this does a check against four tokens, 
        # which is invalid if implemented in RASP,
        # but can be done as series of pairwise comparisons
        output = detokenize(cur_state)
        edges = [(output[i], output[i+1], output[i+2], output[i+3]) 
                 for i in range(g_start, len(output), 4)]
        is_valid = False
        for edge in edges:
            has_t = any([edge[i] == 't' for i in range(4)])
            has_initial = any([edge[i] == correct_initial_node 
                               for i in range(4)])
            if has_t and has_initial:
                is_valid = True
                break
        assert is_valid
        return cur_state
\end{lstlisting}
\end{figure*}

\begin{figure*}[t]
\begin{lstlisting}[language=Python, label=func:non-causal-arms-propagate-backward-targets, caption=RASP program for non-causal encoders arm-wise permutation with $\mathcal{O}(1)$ required layers.]

    def non_causal_arms_propagate_backward_targets(seq, idx, 
                                                   masked_x, g_start, 
                                                   i_nodes, j_nodes):

        # is final if only once in masked_x
        counts = sel_width(select(masked_x, masked_x, equals, causal=False))
        is_final = equals(counts, full(counts, 1))

        # positions of final_nodes nodes in i_nodes
        final_nodes = where(is_final, j_nodes, full(idx, -99)) 
        final_idx = where(is_final, idx, full(idx, -99))
        # positions of final_nodes nodes in i_nodes
        final_idx_slash = where(is_final, idx + 1, full(idx, -99)) 
        is_final_slash = kqv(idx + 1, idx, is_final, equals, 
                             default=False, causal=False)  # shift right
                             
        leading_offset = (arm_len - 3) * 3 + 1
        leading_idx = final_idx - leading_offset
        # gather leading nodes and place in final slash
        leading_nodes = kqv(idx, leading_idx, i_nodes, equals, 
                            default=-99, causal=False)
        leading_nodes = kqv(final_idx_slash, idx, leading_nodes, equals, 
                            default=-99, causal=False)  # shift
        cur_state = where(is_final_slash, leading_nodes, seq)

        # check valid, assume q before g
        output = detokenize(cur_state)
        edges = [(output[i], output[i+1], 
                 output[i+2]) for i in range(g_start, len(output), 3)]
        is_valid = False
        for edge in edges:
            if edge[1] == 't' and edge[2] == correct_initial_node:
                is_valid = True
                break

        assert is_valid
        return cur_state
\end{lstlisting}
\end{figure*}

\begin{figure*}[t]
\begin{lstlisting}[language=Python, label=func:causal-propagate, caption=RASP program for causal encoders/decoders.]
def causal_propagate(seq, idx, masked_x, g_start, i_nodes, j_nodes):
    is_start = equals(i_nodes, full(j_nodes, d['s']))  # get start indices
    # positions of start nodes in i_nodes, requires Q is before G
    # pos of start nodes in i_nodes
    start_idx = where(is_start, idx, full(idx, -99))  
    # find leading nodes and copy leading token to '/'
    leading_idx = kqv(start_idx + 2, idx, idx + 1, equals, 
                      default=-99, causal=True)  # shift
    leading_nodes = kqv(idx, leading_idx, j_nodes, equals, 
                        default=-99, causal=True)

    cur_state = where(is_true(leading_nodes), leading_nodes, seq)

    # edge is (i, j, k) where k started as '/' token and is used for memory
    cur_j_idx = kqv(start_idx + 1, idx, idx + 1, equals, 
                    default=-99, causal=True)  # shift by 1, add 1 to pos
    cur_k_nodes = leading_nodes
    cur_j_nodes = where(is_true(cur_j_idx), j_nodes, full(j_nodes, -99))

    for step in range(0, arm_len - 1):
        # rule 1: connecting edge is before current edge,
        # i.e. cur_j_idx > connecting_i_idx
        # then find match cur_j_node to connecting_i_node 
        # and move connecting_j_node to cur_j_node

        # select width works as binary mask as only one possible match
        is_before = sel_width(select(i_nodes, cur_j_nodes, 
                                     equals, causal=True))  # values at j pos
        connecting_i_idx = kqv(i_nodes, cur_j_nodes, idx,  # values at j pos 
                               equals, default=-99, causal=True)  
        connecting_i_node = kqv(idx - 1, connecting_i_idx, j_nodes, equals, 
                                default=-99, causal=True)  # at j pos
        # update state, only j changes
        cur_j_nodes = where(is_before, connecting_i_node, cur_j_nodes)
        # write in connecting token to k-node pos
        cur_state = where(is_before, connecting_i_node, cur_state)  

        # rule 2: connecting edge is after current edge, 
        # i.e. cur_j_idx < connecting_i_idx
        # find match of cur_j_node to connecting_i_node 
        # and move cur_j_node to connecting_k_node

        # notice the reverse order of the arguments
        is_after = sel_width(select(cur_j_nodes, i_nodes, 
                                    equals, causal=True))
        # at connecting i pos, points to the cur j node pos
        connecting_i_idx = kqv(cur_j_nodes, i_nodes, idx, equals, 
                               default=-99, causal=True)
        # at connecting i pos, points to the cur j node pos
        connecting_k_idx = kqv(idx + 2, idx, connecting_i_idx, equals, 
                               default=-99, causal=True)
        connecting_k_node = kqv(idx, connecting_k_idx + 1, cur_k_nodes, equals, 
                                default=-99, causal=True)
        # update state by moving current edge to future edge
        new_after = is_true(connecting_k_node)  # at k pos
        cur_k_nodes = where(new_after, connecting_k_node, cur_k_nodes)

        cur_j_idx_at_i = where(is_after, idx + 1, full(idx, -99))
        cur_j_idx = kqv(cur_j_idx_at_i, idx, idx + 1, equals, 
                       default=-99, causal=True)  # shift by 1
        new_cur_j_nodes = where(is_true(cur_j_idx), j_nodes, full(j_nodes, -99))
        cur_j_nodes = where(is_true(new_cur_j_nodes), new_cur_j_nodes, cur_j_nodes)
        # write in connecting token to k-node pos
        cur_state = where(new_after, cur_k_nodes, cur_state)  

    # valid if edge[1] == 't' and edge[2] == correct_initial_node
    return cur_state
\end{lstlisting}
\end{figure*}


\subsection{RASP core and library functions}\label{appx:rasp-lib}

We define the required core functions of RASP and helper library functions used in the above algorithms.

\begin{lstlisting}[language=Python, caption=An array filled with `const' of shape `x'.]
def full(x, const):
    return np.full_like(x, const))
\end{lstlisting}

\begin{lstlisting}[language=Python, label=func:inidices, caption=An array of 0 ... len(X).  These correspond to positional embeddings and allow us to make position-wise decisions.]
def indices(x):
    return np.arange(len(x), dtype=int)
\end{lstlisting}

\begin{lstlisting}[language=Python, caption=A bool array.]
def is_true(x, default=0):
    return x > default
\end{lstlisting}

\begin{lstlisting}[language=Python, caption=A bool array.  It is also an example predicate function given to `select'.]
def equals(x, y):
    return x == y
\end{lstlisting}

\begin{lstlisting}[language=Python, caption=An len(k) $\times$ len(k) bool matrix.]
def select(k, q, pred, causal=True):
    s = len(k)
    A = np.zeros((s, s), dtype=bool)
    for qi in range(s):
        # k index <= q index if causal
        for kj in (range(qi + 1) 
            if causal else range(s)):  
            A[qi,kj] = pred(k[kj],q[qi])
\end{lstlisting}

\begin{lstlisting}[language=Python, 
caption=An array counting the number of True values in each row of selection matrix A.]
def sel_width(A):
    o = np.ones(len(A))
    return np.dot(A, o).astype(int)
\end{lstlisting}

\begin{lstlisting}[language=Python, caption=An array ]
def aggr_mean(A, v, default=0):
    out = np.dot(A, v)
    norm = sel_width(A)
    o = np.full_like(v, default, 
                     dtype=float)
    out = np.divide(out, norm, out=o, 
                    where=(norm != 0))
    return out.astype(int)
\end{lstlisting}

\begin{lstlisting}[language=Python, caption=An array. The workhorse of RASP as it mimics the attention mechanism in a transformer.]
def kqv(k, q, v, pred, 
        default=0, causal=True):
    A = select(k, q, pred, causal)
    return aggr_mean(A, v, 
                     default=default)
\end{lstlisting}

\begin{lstlisting}[language=Python, caption=An array.]
def seq_map(x, y, func):
    l = [func(xi, yi) 
         for xi, yi in zip(x, y)]
    return np.array(l).astype(int)
\end{lstlisting}

\begin{lstlisting}[language=Python, caption=An array.]
def where(condition, x_if, y_else):
    x_m = seq_map(x_if, condition, 
                  lambda x, m: 
                    x if m else 0)
    y_m = seq_map(y_else, condition, 
                  lambda y, m: 
                    y if not m else 0)
    return seq_map(x_m, y_m, 
                   lambda x, y: 
                    x if y == 0 else y)
\end{lstlisting}

\end{document}